\documentclass[10pt,journal,twocolumn,twoside]{IEEEtran}
\usepackage{graphicx}
\usepackage{overpic}
\usepackage{enumerate}
\usepackage{multirow}
\usepackage{booktabs}
\usepackage{amsfonts}

\usepackage{threeparttable}
\usepackage{subfigure}
\usepackage{bbding}
\usepackage{cite}
\usepackage[linesnumbered, ruled]{algorithm2e}
\makeatletter
\newcommand{\removelatexerror}{\let\@latex@error\@gobble}
\makeatother
\usepackage{amsmath}
\usepackage{algpseudocode}
\newcommand{\etal}{\textit{et al}. }

\usepackage{color}
\usepackage{url}
\usepackage{amssymb}
\usepackage[normalem]{ulem}
\useunder{\uline}{\ul}{}

\usepackage[breaklinks,colorlinks]{hyperref}

\usepackage[capitalize]{cleveref}
\crefname{section}{Sec.}{Secs.}
\Crefname{section}{Section}{Sections}
\Crefname{table}{Table}{Tables}
\crefname{table}{Tab.}{Tabs.}
\begin{document}

\makeatletter
\DeclareRobustCommand\onedot{\futurelet\@let@token\@onedot}
\def\@onedot{\ifx\@let@token.\else.\null\fi\xspace}

\def\eg{\emph{e.g}\onedot} \def\Eg{\emph{E.g}\onedot}
\def\ie{\emph{i.e}\onedot} \def\Ie{\emph{I.e}\onedot}
\def\cf{\emph{cf}\onedot} \def\Cf{\emph{Cf}\onedot}
\def\etc{\emph{etc}\onedot} \def\vs{\emph{vs}\onedot}
\def\wrt{w.r.t\onedot} \def\dof{d.o.f\onedot}
\def\iid{i.i.d\onedot} \def\wolog{w.l.o.g\onedot}
\def\etal{\emph{et al}\onedot}
\makeatother

\title{Dense Pixel-to-Pixel Harmonization via Continuous Image Representation}
\author{Jianqi~Chen, Yilan~Zhang, Zhengxia~Zou, Keyan~Chen,~and~Zhenwei~Shi*,~\IEEEmembership{Member,~IEEE}
\thanks{The work was supported by the National Key Research and Development Program of China (Grant No. 2022ZD0160401), the National Natural Science Foundation of China under the Grants 62125102, the Beijing Natural Science Foundation under Grant JL23005, and the Fundamental Research Funds for the Central Universities. \textit{(Corresponding author: Zhenwei Shi (email: shizhenwei@buaa.edu.cn))}}%
\thanks{Jianqi Chen, Yilan Zhang, Keyan Chen, and Zhenwei Shi are with the Image Processing Center, School of Astronautics, Beihang University, Beijing 100191, China, and with the State Key Laboratory of Virtual Reality Technology and Systems, Beihang University, Beijing 100191, China, and also with the Shanghai Artificial Intelligence Laboratory, Shanghai 200232, China.}
\thanks{Zhengxia Zou is with the Department of Guidance, Navigation and Control, School of Astronautics, Beihang University, Beijing 100191, China.}
\thanks{Copyright © 2023 IEEE. Personal use of this material is permitted. However, permission to use this material for any other purposes must be obtained from the IEEE by sending an email to pubs-permissions@ieee.org.}
}

\markboth{IEEE TRANSACTIONS ON CIRCUITS AND SYSTEMS FOR VIDEO TECHNOLOGY,~Vol.~XX, No.~X, August~XXXX}%
{Chen \MakeLowercase{\textit{et al.}}: Dense Pixel-to-Pixel Harmonization via Continuous Image Representation}

\maketitle

\begin{abstract}
High-resolution (HR) image harmonization is of great significance in real-world applications such as image synthesis and image editing. However, due to the high memory costs, existing dense pixel-to-pixel harmonization methods are mainly focusing on processing low-resolution (LR) images. Some recent works resort to combining with color-to-color transformations but are either limited to certain resolutions or heavily depend on hand-crafted image filters. In this work, we explore leveraging the implicit neural representation (INR) and propose a novel image Harmonization method based on Implicit neural Networks (HINet), which to the best of our knowledge, is the first dense pixel-to-pixel method applicable to HR images without any hand-crafted filter design.  Inspired by the Retinex theory, we decouple the MLPs into two parts to respectively capture the content and environment of composite images. A Low-Resolution Image Prior (LRIP) network is designed to alleviate the Boundary Inconsistency problem, and we also propose new designs for the training and inference process. Extensive experiments have demonstrated the effectiveness of our method compared with state-of-the-art methods. Furthermore, some interesting and practical applications of the proposed method are explored. Our code is available at \href{https://github.com/WindVChen/INR-Harmonization}{https://github.com/WindVChen/INR-Harmonization}. 

\end{abstract}


\IEEEpeerreviewmaketitle

\begin{figure}[t]
  \centering
   \includegraphics[width=\linewidth]{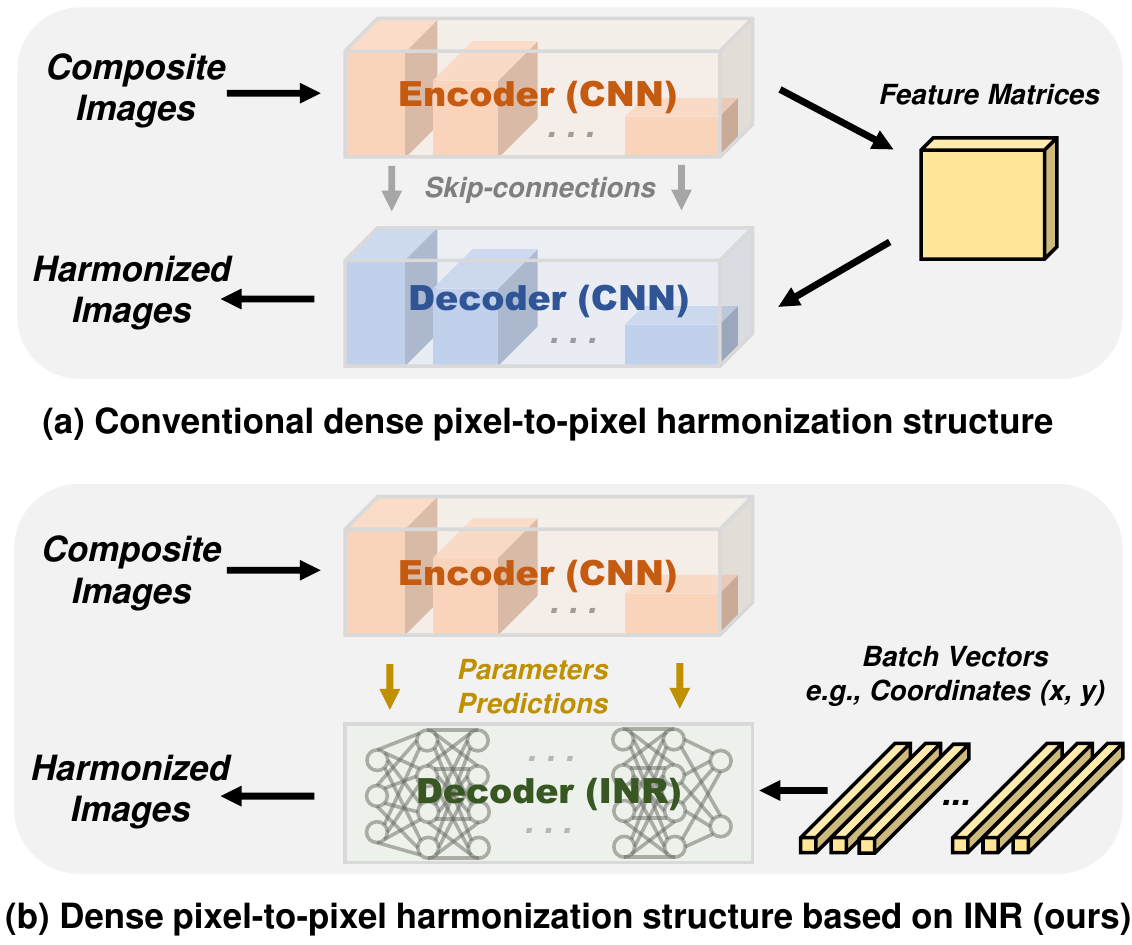}
   \caption{Structural differences between the existing harmonization methods and our proposed HINet.}
   \label{fig:Diff Decoder}
\end{figure}

\section{Introduction}
\label{sec:intro}

\IEEEPARstart{I}{mage} compositing, a fundamental technique in image processing, has been widely used in various applications such as image editing\cite{10.1145/1201775.882269, barnes2009patchmatch, kwatra2003graphcut}, data augmentation\cite{yun2019cutmix, zhang2020learning, wang2020constrained}, \etc. It encompasses various methods such as image matting \cite{wang2023composited} and shadow generation/removal \cite{inoue2020learning}, with the objective of generating realistic synthetic outputs by extracting foreground objects from one image and seamlessly integrating them into another background image. Nonetheless, inconsistencies in color spaces between the foreground and background of the composite image often result in perceptual disparities due to variations in lighting and tonal qualities. This challenge, distinct from image enhancement techniques \cite{ren2018lecarm, zhao2021retinexdip} which primarily address the overall appearance of an image, frequently necessitates manual adjustments of the color distribution within the foreground image layers, which demands much professional knowledge and does cost lots of time.

Aimed to harmonize the composite images, \cite{tsai2017deep} first introduced deep learning into the task and many effective approaches have been proposed \cite{cong2020dovenet, guo2021image, ling2021region, guo2021intrinsic, sofiiuk2021foreground, hang2022scs} in recent years. These data-driven approaches, compared with traditional ones \cite{lalonde2007using, xue2012understanding, reinhard2001color, pitie2005n} that rely on matching low-level statistics between foreground and background, have demonstrated better results with their strong semantic representation capability. However, as these methods mostly adopt a U-Net \cite{ronneberger2015u} like structure, the harmonization process is essentially a dense pixel-to-pixel transformation and costs much GPU memory \cite{cong2022high}. As a result, it is difficult for these methods to be applied for processing high-resolution (HR) images, such as $2K$ or $4K$, and most of them only perform at a low resolution of 256$\times$256 pixels. To achieve HR image harmonization, more recent works \cite{cong2022high, ke2022harmonizer, xue2022dccf} proposed to leverage color-to-color transformations which saves much memory. Despite the applicability of harmonizing HR images, these methods are either limited to certain resolutions \cite{cong2022high}, or heavily rely on hand-crafted filters \cite{ke2022harmonizer, xue2022dccf} which are cumbersome in design and also limit the potential of deep learning networks.

Considering the unaffordable memory cost of the current pixel-to-pixel deep learning-based harmonization frameworks, we study if it is possible to apply a scale-adaptive dense pixel-to-pixel transformation to HR image harmonization. In our method, we are inspired by the recently popular paradigm Implicit Neural Representation (INR) \cite{sitzmann2020implicit, tancik2020fourier}, which leverages a stack of multilayer perceptrons (MLP) to represent a 3D scene \cite{mildenhall2021nerf} or a 2D image \cite{sitzmann2020implicit} parameterized by continuous coordinate input. Two features of INR appeal to us. Firstly, different from the CNN structure that takes the whole image/feature map as input, the input to INR structure is a vector containing grid coordinates, providing us more control over the memory cost. Secondly, by inputting continuous coordinates $(x, y)$ and outputting RGB values, the MLP models in its weights a continuous image that is not limited by resolution, which may benefit HR image harmonization. Inspired by these two advantages, in this paper we explore a novel dense pixel-to-pixel image Harmonization method based on Implicit neural Networks (HINet). \cref{fig:Diff Decoder} shows the differences between our method and previous harmonization methods.

Directly applying conventional INR approaches to the image harmonization task can lead to several issues, including subpar performance, discontinuous patterns, and high memory consumption. We have provided an extensive analysis of these challenges in \cref{subsec:analysis}. In light of these challenges, we have meticulously designed the proposed HINet to achieve superior harmonization results. Inspired by Retinex theory\cite{land1971lightness, land1977retinex}, we decouple the MLPs in the HINet decoder into a content extraction part and an appearance rendering part. One is to preserve content structure and determine what objects are in the image, while the other is to capture the global environment of the image. The parameters of these two parts' MLPs are predicted by different encoder layers, thus alleviating the burden of the last output layer. Furthermore, to prevent the potential inconsistency problem of the local MLPs, we design a Low-Resolution Image Prior (LRIP) structure, where the MLPs of the content extraction part are more finely divided into several parts, each processing a specific resolution of the input with the former lower resolution part's output as a prior. This structure can both solve the inconsistency problem and reduce memory costs. For HR image harmonization training, we design a Random Step Crop (RSC) strategy, while also introducing the inference process for ultra-HR images. Moreover, following \cite{ke2022harmonizer, xue2022dccf}, we include an optional 3D lookup table (LUT)\cite{karaimer2016software, zeng2020learning} rendering branch to enrich the comprehensibility of the model and enable the manual control.

Our contributions can be summarized as follows:

\begin{itemize}
    \item As far as we know, our method is the first dense pixel-to-pixel image harmonization method that can be applied to HR images. The proposed HINet requires no hand-crafted filters and liberates the strong learning ability of the neural networks.
    \item We explore leveraging INR in image harmonization. We expect that the new paradigm can pave way for more future research on HR harmonization task.
    \item Extensive experiments have demonstrated the effectiveness of our method. Compared with previous methods, the HINet can achieve state-of-the-art performance on HR image harmonization. Moreover, we have explored some interesting potentials of the HINet in practical usages, such as arbitrary resolution harmonization, and region-based harmonization for both images and videos.
\end{itemize}

\begin{figure*}[t]
  \centering
   \includegraphics[width=\textwidth]{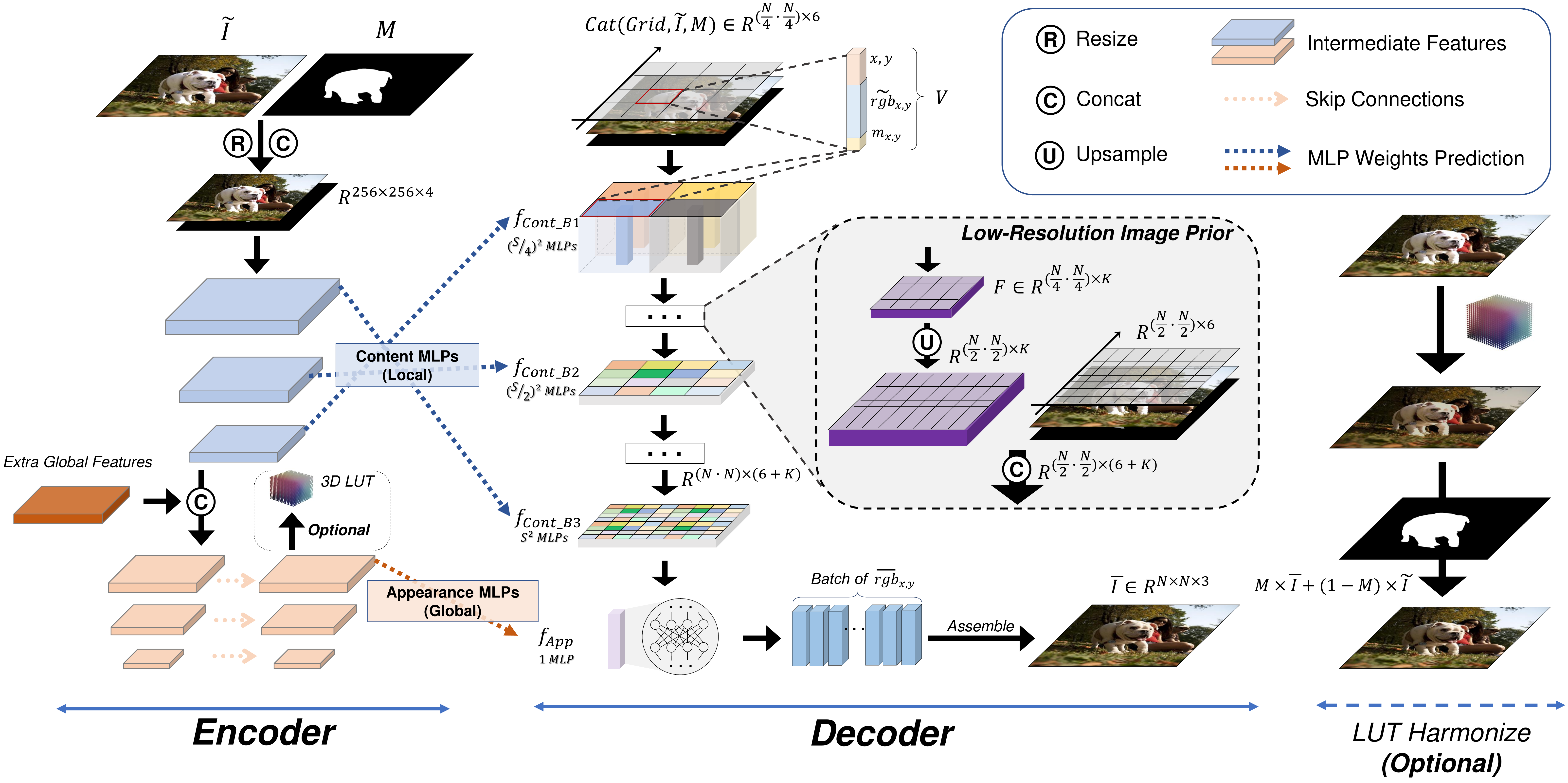}
   \vspace{-3ex}
   \caption{The pipeline of our method. The HINet consists of an Encoder, a Decoder, and an optional LUT Harmonize module. Given a downsampled composite image $\widetilde{I}$ and its mask $M$, the Encoder predicts parameters of the decoder's MLPs and 3D LUT (optional). Fix the MLPs' parameters, we feed into the decoder a batch of vectors $V$, which is a concatenation of grid coordinate $(x, y)$, value $m_{x, y}$ in $M$, and value $\widetilde{rgb}_{x, y}$ in $\widetilde{I}$. We then assemble the output vectors $\overline{rgb}_{x, y}$, and obtain harmonized images $\overline{I}$. Note that the number of layers in the figure is simplified, please refer to \cref{subsec:exp settings} for more details. Details of the Encoder structure can be referred to \cite{sofiiuk2021foreground}, where ``Extra Global Features" denotes the features from an additional HRNet \cite{wang2020deep}.}
   \label{fig:Pipeline}
\end{figure*}

\section{Related Works}
\label{sec:relate}

\textbf{Image Harmonization.} In this part, we mainly focus on reviewing deep learning-based image harmonization methods, as their superiority over the traditional methods\cite{lalonde2007using, xue2012understanding, reinhard2001color, pitie2005n, sunkavalli2010multi} has been demonstrated \cite{tsai2017deep}.  Since Tsai \etal \cite{tsai2017deep} pioneered conducting image harmonization with deep neural networks,  many data-driven methods\cite{cun2020improving, cong2020dovenet, guo2021image, ling2021region, sofiiuk2021foreground, hang2022scs} have been proposed and achieve good results on LR images. Cun \etal \cite{cun2020improving} proposed to leverage the attention mechanism for better learning features of foreground and background. Cong \etal \cite{cong2020dovenet} considered the domain shift to harmonize composited images. Ling \etal \cite{ling2021region} referred to AdaIN\cite{huang2017arbitrary} and regarded image harmonization as a style transfer problem. Sofiiuk \etal \cite{sofiiuk2021foreground} proposed to utilize high-level semantic features from pre-trained models. Hang \etal \cite{hang2022scs} leveraged contrastive learning to narrow the solution space.  These methods, although performing well on LR image harmonization ($256\times256$), are hard to be applied to harmonize HR images due to the high memory cost in their U-Net\cite{ronneberger2015u} structure design.  To meet the practical needs of HR image harmonization, more recent works \cite{cong2022high, ke2022harmonizer, xue2022dccf, liang2022spatial, wang2023semi, guerreiro2023pct} resorted to color-to-color operations. Cong \etal \cite{cong2022high} proposed to combine features of pixel-to-pixel and color-to-color transformations, and \cite{ke2022harmonizer, xue2022dccf} manually designed several image filters and predicted their parameters. Despite their potential in processing HR images, there remain some limitations. For CDTNet \cite{cong2022high}, since the Refinement Module alone requires about 6GB memory for a single 2048$\times$2048 image, it will consume unbearable memory at higher resolution (\eg 6K) and thus only works for certain high resolutions. For \cite{ke2022harmonizer, xue2022dccf}, although they are capable of being applied to flexible high resolutions, these methods rely heavily on hand-crafted filters that are cumbersome in design \cite{ke2022harmonizer, xue2022dccf}. Recent \cite{liang2022spatial} proposed to approximate the filters with predicted piece-wise linear functions, yet may fail to model complex scenes due to the simplicity of the function. Another more recent study \cite{wang2023semi} also utilizes piece-wise linear curves and predicts additional shading maps to enhance local control. Additionally, \cite{guerreiro2023pct} approximates color transformations through the application of an affine matrix and predicts a corresponding parameter map.

In contrast to previous dense pixel-to-pixel harmonization approaches relying on CNN structures, our work introduces an INR-based method meticulously tailored for processing high-resolution composite images in a dense pixel-to-pixel manner. Notably, it represents the first dense pixel-to-pixel high-resolution image harmonization method. The utilization of this dense pixel-to-pixel approach enables us to harness the full potential of deep networks, surpassing the capabilities of hand-crafted filters commonly employed in color-to-color methods. This enhancement empowers us to effectively address more complex scenarios, ultimately yielding state-of-the-art performance.

\textbf{Implicit Neural Representation.} The INR method was originally proposed in \cite{stanley2007compositional}, and has gained much popularity in 3D area recently\cite{mildenhall2021nerf, park2019deepsdf, mescheder2019occupancy}, where it can represent a continuous 3D shape and is a memory-economic way compared to traditional approaches\cite{choy20163d, fan2017point, wang2018pixel2mesh, wang2021lossy, benedek2016lidar} such as point cloud and voxel. The key idea of INR is to convert originally sparse coordinates into continuous signals. Some recent studies  \cite{tancik2020fourier,sitzmann2020implicit} show that INR with Fourier embedding and periodic activation like sinusoidal can be well applied to 2D area and represent photorealistic images. Since the coordinates are in a continuous real space, the generated images are then continuous. Inspired by INR, many recent works have explored leveraging it into different tasks and achieving some good results, \eg image-to-image translation\cite{shaham2021spatially}, image super-resolution\cite{chen2021learning}, image generation\cite{anokhin2021image, skorokhodov2021adversarial, karras2021alias}, \etc. Different from these works, we focus on building a dense pixel-to-pixel image harmonization network that can be applied to ultra-HR images. The HINet structure has been carefully designed and some interesting potentials for practical use have been explored.

\section{Proposed Method}
\label{sec:method}

\subsection{Overview}
\label{subsec:overview}

The HINet architecture consists of an encoder and a decoder. The encoder structure aligns with prior research \cite{sofiiuk2021foreground, xue2022dccf}, incorporating additional global features within its intermediate layer. In the decoder segment, we adhere to the INR paradigm, employing a stack of MLPs. The overall architecture is illustrated in \cref{fig:Pipeline}. Given a composite image $\widetilde{I}$, we input its resized version (256x256) and the corresponding mask $M$ into the encoder. This step enables the prediction of parameters for the decoder MLPs. Once the decoder's weights are determined, we feed it with a batch of vectors $V$, with a batch size matching the pixel count of the original composite image. The vector $V$ represents a concatenation of grid coordinates $(x, y)$, mask values $m_{x, y}$, and composite image values $\widetilde{rgb}_{x, y}$. These vectors undergo processing by the decoder MLPs, yielding harmonized signals $\overline{rgb}_{x, y}$. By assembling these output RGB values, we generate the final harmonized image $\overline{I}$.

In the subsequent subsections, we will delineate the challenges associated with implementing INR for image harmonization in Section B. Following this, Sections C through E will provide an in-depth exposition of our network designs, specifically crafted to address these challenges.

\subsection{Analysis of Existing Challenges}
\label{subsec:analysis}

Leveraging INR for the image harmonization task is a challenging task. Recently, there have been many approaches\cite{anokhin2021image, skorokhodov2021adversarial, karras2021alias, shaham2021spatially} applying INR to tasks like image generation and image translation. These methods usually take encoder's output features as the MLPs' weights and get generated images by feeding in coordinates. Some of them apply a stack of globally representative MLPs \cite{anokhin2021image} where every coordinate is processed by the same MLP, while others apply locally representative MLPs\cite{shaham2021spatially} that coordinates are split into different parts and processed by corresponding MLPs. Although these methods achieved attractive results in some tasks, there are mainly three challenges in transferring to image harmonization task.

The first challenge is the design of the INR structure. Since the harmonization task requires preserving content structure while aligning color space between foreground and background, we may encounter content loss \cite{shaham2021spatially} if we choose to utilize global MLPs. Whereas if we choose to leverage local MLPs, the memory cost will dramatically increase which is unaffordable, especially for real harmonization scenarios where images can reach high resolutions, and it may also introduce the problem of inconsistency in the boundaries of adjacent image regions (see \cref{subsec:LRIP}).

The second challenge is the insufficiency of the encoder output features. Existing methods mostly predict the parameters of all MLPs only from the features of the last encoder layer. Although deep layers do capture rich semantic information, content structure information is not well preserved, and it is burdensome to predict a large number of parameters by a single layer. A downsampling operation may reduce the number of parameters, yet inevitably decrease the output image fidelity due to more structural information loss.

The third challenge is how to perform HR training and inference without consuming too much memory. Real-world harmonization scenarios often encounter ultra-HR images, \eg, $6K$. Even just harmonizing a single one (\eg, $6048 \times 4032$ pixels) will consume lots of memory as the vectors input to INR can build a huge batch ($\approx10^7$). Dealing with such an ultra-HR problem is still underexplored by previous INR works.

Considering the aforementioned challenges, we have meticulously crafted HINet. These challenges are individually tackled by our solutions in \cref{subsec:decoupled}, \cref{subsec:LRIP}, and \cref{subsec:specific}.

\subsection{Decoupled Content and Appearance MLPs}
\label{subsec:decoupled}

Existing INR approaches either leverage a stack of globally representative MLPs \cite{skorokhodov2021adversarial, anokhin2021image} where each layer is a single MLP, or locally representative MLPs \cite{shaham2021spatially} where each layer is an MLP matrix that consists several MLPs (The differences are displayed in \cref{fig:Pipeline}). The former may have a low fidelity problem \cite{sitzmann2020implicit}, while the latter may have an out-of-memory problem as there are more MLPs.

Referring to Retinex theory \cite{land1971lightness, land1977retinex} that decomposes an image into illumination and reflectance, we can also decompose the harmonization task into two pieces: determining the environment and the content objects of an image. Along with this idea, we decouple the MLPs in the decoder into a content extraction part $f_{Cont}$ and an appearance rendering part $f_{App}$. Specifically, $f_{Cont}$ leverages locally representative MLPs to both extract objects information and ensure content structure retention, while $f_{App}$ adopts a globally representative MLP to capture the background environmental condition. Since the content structure is mostly preserved in low-level features and the objects recognition only requires local receptive fields, we predict the parameters of $f_{Cont}$ by shallow encoder layers. For $f_{App}$, we predict its parameters from deep layers that can capture global and high-level features, thus conducive to environment capture.

It should be noted that recent works \cite{guo2021intrinsic, guo2021image} also build networks from the perspective of Retinex. However, unlike their strict adherence to Retinex theory (explicitly output illumination and reflectance images, and then multiplying the two), we implicitly embed the idea into the decoder structure design to extract content and environment information. To the best of our knowledge, our design of decoupled MLPs has not been previously explored by other INR works. Moreover, such a design not only offloads the last layer of the encoder and makes the structure of the encoder well aligned with that of the decoder, but also benefits from both local and global MLPs (as shown in \cref{tab:Decouple}).

\begin{figure}[t]
  \centering
   \includegraphics[width=0.8\linewidth]{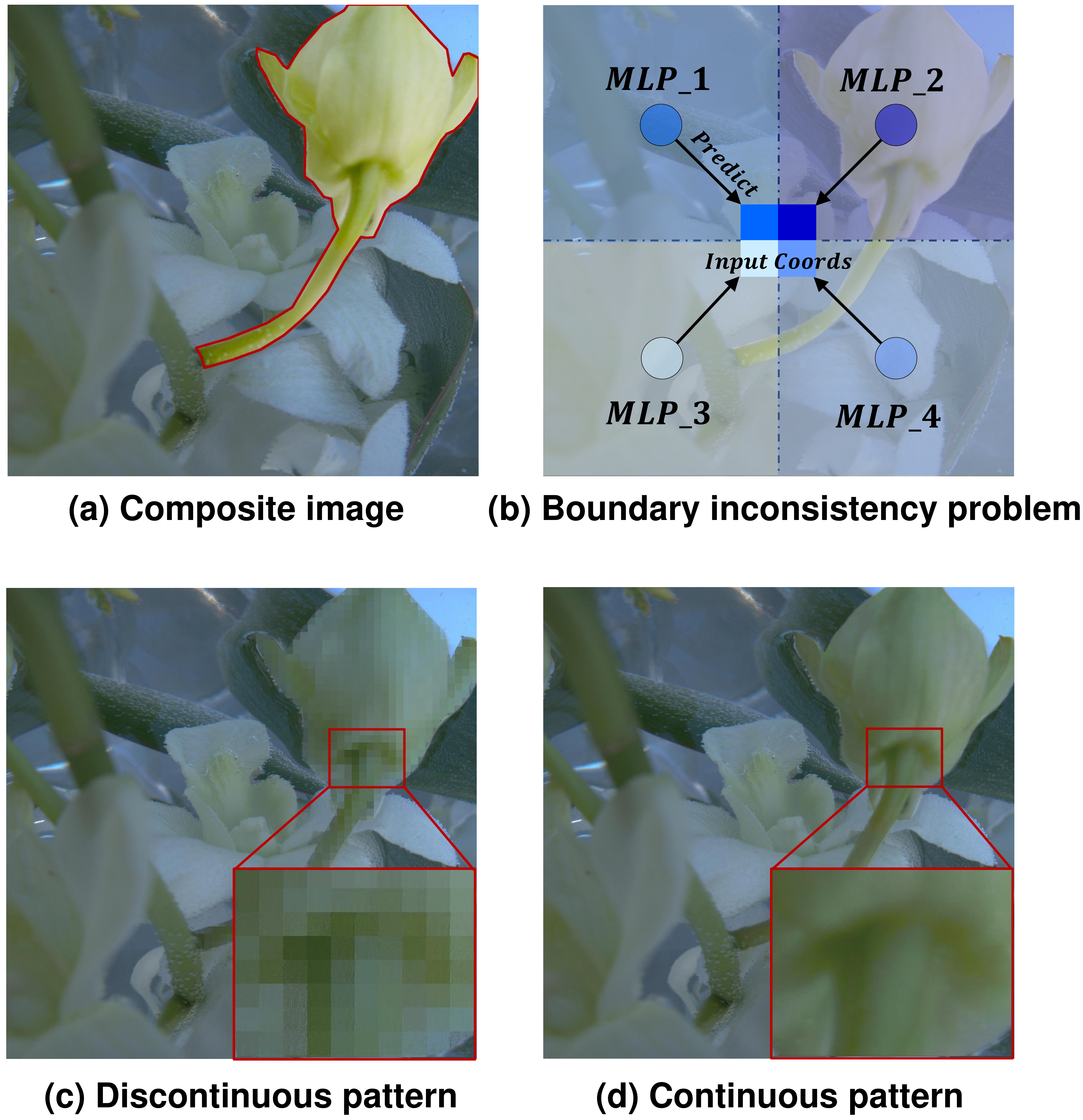}
   \caption{Boundary Inconsistency Problem Illustration. When utilizing locally representative MLPs on a composite image (a), the inconsistency problem emerges. To clearly illustrate this, we consider only four local MLPs, as depicted in (b). Near the boundary of these MLPs, the MLP processing the input coordinate abruptly transitions from one to another, causing a discontinuous pattern, as seen in (c). In contrast, our designed LRIP structure ensures a continuous result, as depicted in (d).}
   \label{fig:Edge_Inconsistency}
\end{figure}

\subsection{Low-Resolution Image Prior}
\label{subsec:LRIP}

In this section, we first discuss the boundary inconsistency problem caused by the design of the $f_{Cont}$, then we introduce our solution. To be specific, since $f_{Cont}$ adopts an MLP matrix structure, the input is divided into several parts, each corresponding to a specific MLP. Therefore, at the boundary of two adjacent MLPs, the processing MLP will suddenly switch from one to another, leading to a discontinuous pattern. We illustrate this problem in \cref{fig:Edge_Inconsistency}.

A very straightforward solution for the above problem is to leverage bilinear interpolation instead of the nearest match. For each input, we query its nearest four corner MLPs and calculate the interpolated MLP. Take the four corner MLPs as $f^p$, where $p \in \{1: 4\}$, from the top-left corner to the bottom-right one, and the area enclosed by the current position and each corner as $s^p$, then the generated MLP can be formulated as:
\begin{equation}
  f_{gen} = \sum_{p\in \{1:4\}}\frac{s^{p'}}{s^{all}}\cdot f^p
  \label{eq:bilinear2}
\end{equation}
where $s^{all}=\sum_{p\in \{1:4\}}s^p$ and $p'$ is the opposite corner of $p$. In this way, each input vector is processed by a continuous MLP matrix, thus alleviating the inconsistency problem. However, although the bilinear interpolation strategy looks simple and effective, it is not feasible in practice. For example, suppose the original MLP matrix is composed of $16\times16$ MLPs, in this case, even if we harmonize an LR $256\times256$ image, then by using the interpolation strategy, we will finally generate a $256\times256$ MLP matrix, an unbearable $\times256$ increase in the number of parameters.

To both reduce the memory cost and avoid the boundary inconsistency between blocks, we propose a new network structure named ``Low-Resolution Image Prior (LRIP)''. Specifically, we divide the MLPs in $f_{Cont}$ into several blocks. We feed the input vectors to each block, and the batch size increases hierarchically. Except for the first one, each block is conditioned on the output features of the previous block (See \cref{fig:Pipeline} for more details). Given a $256\times256$ image and an LRIP structure with two blocks $B_1, B_2$ (for simplicity), the process is defined as follows:
\begin{equation}
  F_1 = B_1(V_{128^2})
  \label{eq:LRIP1}
\end{equation}
\begin{equation}
  F_2 = B_2(Cat(V_{256^2}, Up(F_1)))
  \label{eq:LRIP2}
\end{equation}
where $V_{N^2}$ denotes the input vectors with a batch size of $N^2$, $F$ is the output feature which has the same batch size as the input, $Cat(\cdot)$ is the concatenation operation, $Up(\cdot)$ is the upsampling operation. We adopt the bilinear upsampling in LRIP. In this way, we convert the idea of continuous MLPs to continuous input. Each input is conditioned on the previous block, thus learning a more global representation and can alleviate the inconsistency problem effectively. Furthermore, since the blocks (except the last one) have a lower resolution input, the LRIP structure can save a lot of memory while maintaining high-quality results.

Note that in \cite{skorokhodov2021adversarial}, the authors designed multi-scale INRs which seems similar to our LRIP structure. However, there are many differences. The main difference is in the decoder structure. Due to the different aims, \cite{skorokhodov2021adversarial} merely used a stack of global MLPs as the decoder structure, while in our design, we need to leverage local MLPs to ensure content retention. Furthermore, their decoder's parameters are all from the output features of the encoder's last layer, while ours are well aligned with the encoder structure which can make full use of all encoder layers.

\begin{table*}[t]
\caption{Comparisons with recent state-of-the-art HR harmonization methods \cite{cong2022high, xue2022dccf, ke2022harmonizer}. Since CDTNet \cite{cong2022high} is not open source and only works on certain high resolutions as discussed in \cref{sec:relate}, we directly quote their CDTNet-256 results (not CDTNet-512, whose input configuration is not aligned with other HR methods) on HAdobe5K sub-dataset. For Harmonizer \cite{xue2022dccf} and DCCF \cite{ke2022harmonizer}, as there missed some metric results in the original papers, we re-run their inference code on the original resolution version of the iHarmony4 dataset \cite{cong2020dovenet} with the same device. The best result is shown in bold.}
    \label{tab:SOTA compare}
\centering
    \renewcommand{\arraystretch}{1.2} 
    \centering
    \resizebox{0.43\linewidth}{!}{
\begin{tabular}{c|c|cc}
\hline
HAdobe5K                       & Metric & CDTNet\cite{cong2022high}     & Ours            \\ \hline
\multirow{4}{*}{$1024\times1024$}     & MSE$\downarrow$    & \textbf{21.24}  & 22.68           \\
                          & fMSE$\downarrow$   & \textbf{152.13}   & 187.97          \\
                          & PSNR$\uparrow$   & \textbf{38.77}  & 38.38           \\
                          & SSIM$\uparrow$   & 0.9868         & \textbf{0.9886} \\ \hline
\multirow{4}{*}{$2048\times2048$}     & MSE$\downarrow$    & 29.02          & \textbf{24.08}  \\
                          & fMSE$\downarrow$   & 198.85     & \textbf{192.20}  \\
                          & PSNR$\uparrow$   & 37.66       & \textbf{38.35}  \\
                          & SSIM$\uparrow$   & 0.9845       & \textbf{0.9886} \\ \hline
\multirow{4}{*}{\shortstack{Original resolution \\ ($\sim6K$)}} & MSE$\downarrow$    & -               & \textbf{21.81}  \\
                          & fMSE$\downarrow$   & -           & \textbf{173.72} \\
                          & PSNR$\uparrow$   & -             & \textbf{38.71}  \\
                          & SSIM$\uparrow$   & -              & \textbf{0.9871} \\ \hline
 \end{tabular}}
\hspace{1em}
\tiny
\renewcommand{\arraystretch}{0.8} 
    \resizebox{0.53\linewidth}{!}{
\begin{tabular}{c|c|cc|c}
\hline
Dataset    & Metric & Harmonizer\cite{ke2022harmonizer}      & DCCF\cite{xue2022dccf}            & Ours            \\ \hline
\multirow{4}{*}{HAdobe5K}   & MSE$\downarrow$    & 24.09           & 23.12           & \textbf{21.45}  \\
           & fMSE$\downarrow$   & 193.70          & 195.60          & \textbf{172.79} \\
           & PSNR$\uparrow$   & 37.82           & 37.78           & \textbf{38.67}  \\
           & SSIM$\uparrow$   & 0.9339          & 0.9858          & \textbf{0.9873} \\ \hline
\multirow{4}{*}{HCOCO}      & MSE$\downarrow$    & 20.39           & \textbf{16.84}  & 17.29           \\
           & fMSE$\downarrow$   & 364.52          & 317.43          & \textbf{315.98} \\
           & PSNR$\uparrow$   & 37.80           & \textbf{38.65}  & \textbf{38.65}  \\
           & SSIM$\uparrow$   & 0.9858          & \textbf{0.9929} & 0.9927          \\ \hline
\multirow{4}{*}{Hday2night} & MSE$\downarrow$    & \textbf{37.72}  & 55.78           & 51.24           \\
           & fMSE$\downarrow$   & \textbf{636.04} & 715.52          & 713.66          \\
           & PSNR$\uparrow$   & 37.20           & \textbf{37.52}  & 37.35           \\
           & SSIM$\uparrow$   & 0.9548          & 0.9787          & \textbf{0.9801} \\ \hline
\multirow{4}{*}{Hflickr}    & MSE$\downarrow$    & 67.82           & \textbf{64.62}  & 66.56           \\
           & fMSE$\downarrow$   & 473.30          & \textbf{438.44} & 449.71          \\
           & PSNR$\uparrow$   & 33.44           & \textbf{33.61}  & 33.56           \\
           & SSIM$\uparrow$   & 0.9714          & 0.9843          & \textbf{0.9844} \\ \hline
\multirow{4}{*}{All}        & MSE$\downarrow$    & 27.09           & 24.72           & \textbf{24.62}  \\
           & fMSE$\downarrow$   & 331.73          & 302.57          & \textbf{296.31} \\
           & PSNR$\uparrow$   & 37.31           & 37.84           & \textbf{38.07}  \\
           & SSIM$\uparrow$   & 0.9685          & 0.9896          & \textbf{0.9900} \\ \hline
\end{tabular}}
\end{table*}

\begin{figure*}[!ht]
 \centering
   \begin{minipage}[t]{0.44\linewidth}
  \centering
   \includegraphics[width=\linewidth]{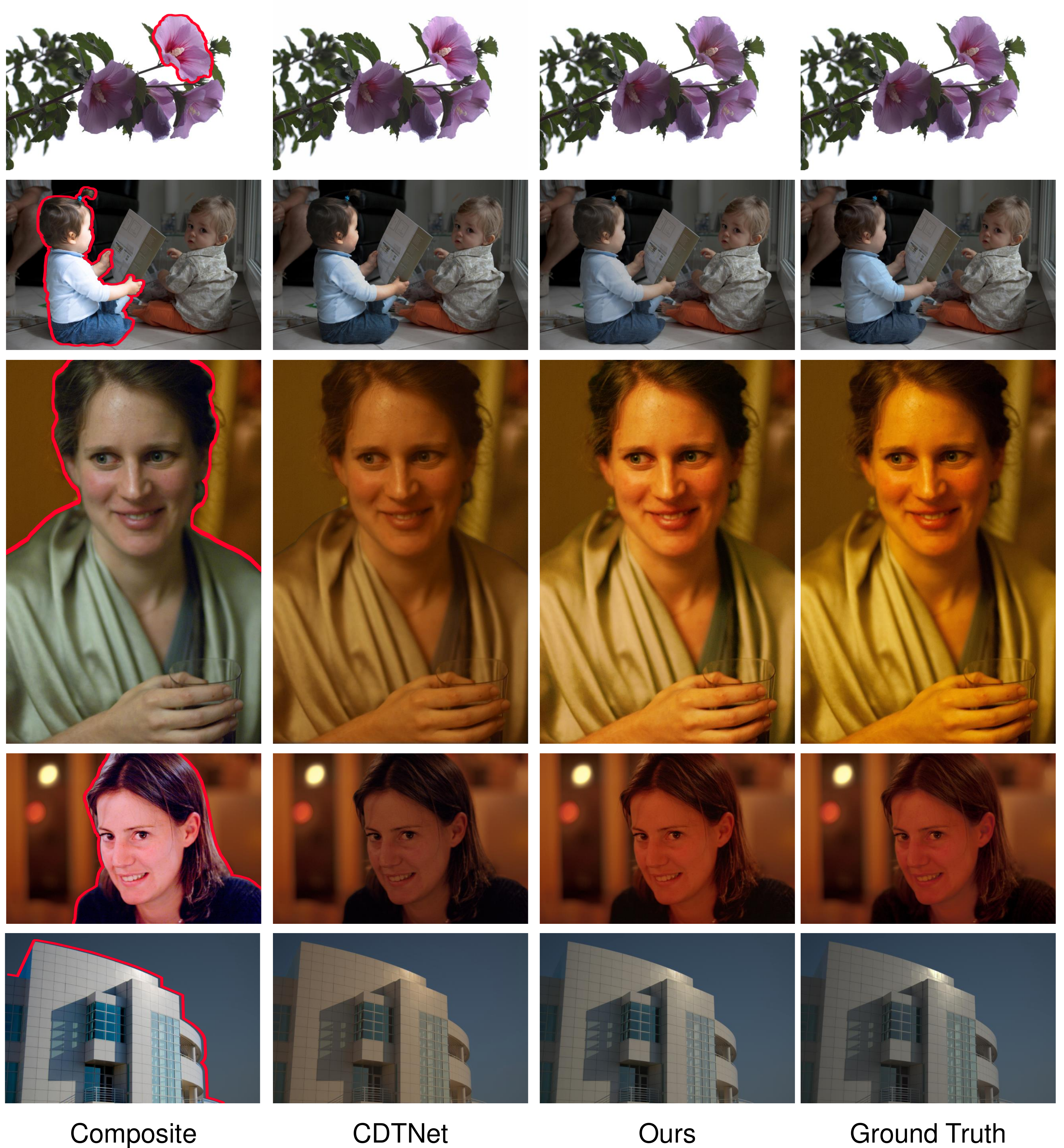}
   \caption{Visual comparisons on $2048\times 2048$ HR version of HAdobe5K sub-dataset. From left to right, we show the composite images, the results of \cite{cong2022high} and ours, and the ground truth images. The foreground is stroked by a red line. We have resized the images to their original aspect ratio for a better view.}
   \label{fig:VISAdobe2048}
\end{minipage}
\hspace{1em}
 \begin{minipage}[t]{0.51\linewidth}
  \centering
   \includegraphics[width=\textwidth]{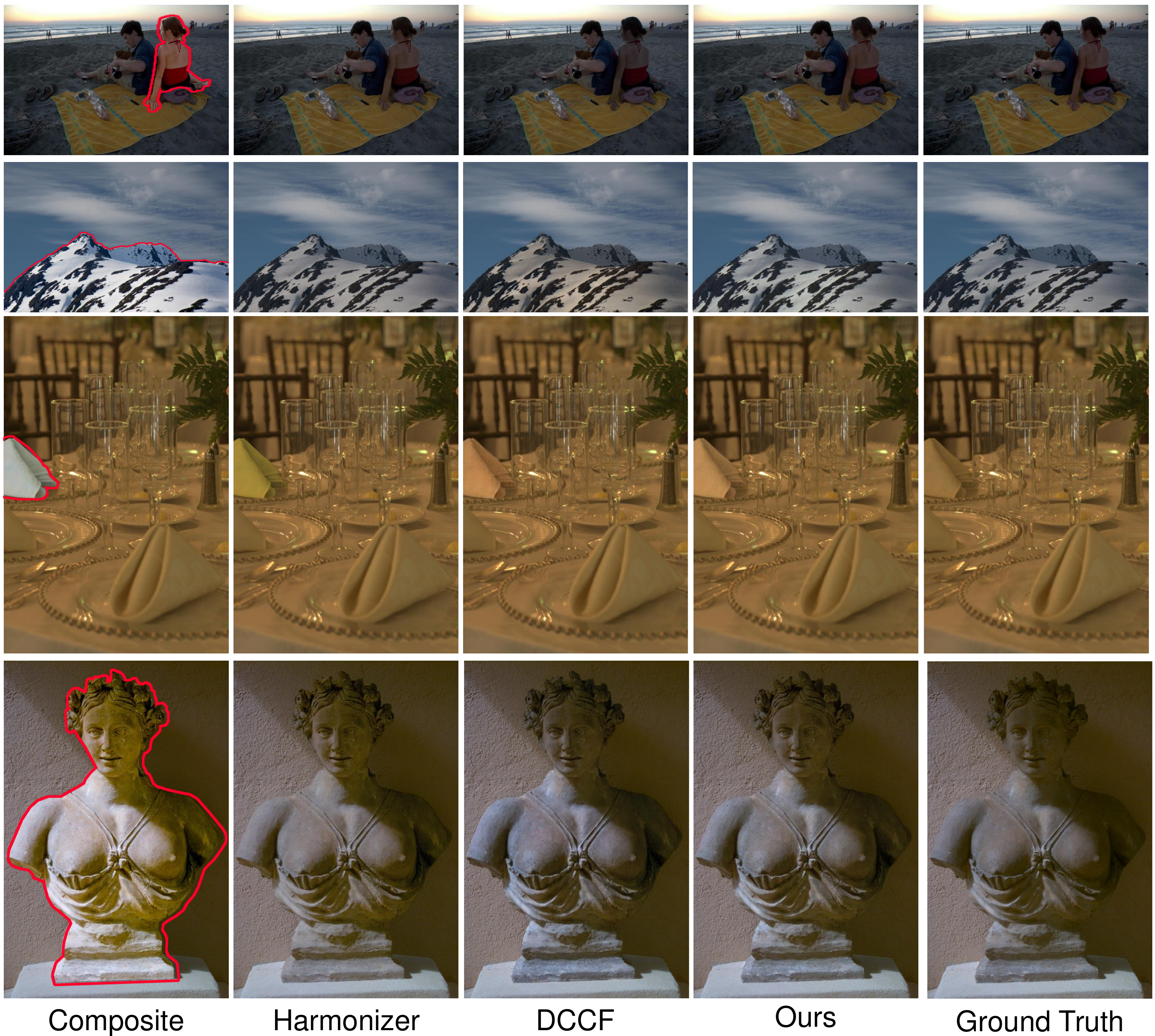}
   \caption{Visual comparisons on the original resolution of iHarmony4 dataset (resolution can reach $6K$). We are the first dense pixel-to-pixel method that can be applied to the original resolution. From left to right, we show the composite images, the results of \cite{ke2022harmonizer, xue2022dccf} and ours, and the ground truth images. The foreground is stroked by a red line. Please zoom in for a better view.}
   \label{fig:VISRAW}
   \end{minipage}
\end{figure*}

\subsection{HR Image Harmonization}
\label{subsec:specific}

Compared with LR image harmonization, it is more challenging to harmonize an HR image, especially for dense pixel-to-pixel transformation methods. We, therefore, propose designs for both the training and inference processes.

\textbf{Multiple inputs.} The conventional input of the INR is the coordinate $(x, y)$ \cite{tancik2020fourier, sitzmann2020implicit, skorokhodov2021adversarial, anokhin2021image}. Although with only the coordinate as input, we can achieve some good results in LR image harmonization, the quality deteriorates sharply when harmonizing higher-resolution images. Considering that the network only sees the LR image (the encoder's input is a down-sampled version of the image), when applied to HR image harmonization, the decoder is required to not only do harmonization but also super-resolution. We show that this will be a much more challenging task and the network just fails to achieve both. Therefore, in practice, apart from the coordinate $(x, y)$, we also feed the composite image's RGB value $\widetilde{rgb}_{x, y}$ and the mask value $m_{x, y}$ into the input vector, which is expected to provide guidance for the decoder and make it focus on the harmonization task. In this way, our input is finally a 6D vector that can be formulated as $V=(x, y, \widetilde{rgb}_{x, y}, m_{x, y})$.

\textbf{HR training process.} When harmonizing LR images, the training process is straightforward and we can just feed all the input vectors into the decoder. However, as mentioned in \cref{sec:intro}, it is unaffordable for HR image harmonization due to the huge batch size. Benefiting from the advantage of the INR that the input is a batch of vectors, rather than a whole feature matrix as CNN, we can feed partial vectors into the decoder, and we design a Random Step Crop (RSC) strategy which is simple but effective. To be specific, we crop out the same local area from the composite image, the coordinate map, and the mask, and feed the vectors in this area into the decoder. The RSC strategy is somewhat similar to the regular RandomCrop augmentation, except that we not only need to crop the original resolution images but also the downsampled ones to meet the needs of the LR input in the LRIP structure (Please refer to \cref{subsec:LRIP}.). Furthermore, the motivation of the RSC strategy is for the feasible HR image training but not the data augmentation. Following \cite{xue2022dccf}, we also employ a progressive training strategy, first training on LR images and then finetuning on HR ones, which can bring in better results.

\textbf{HR inference process.} Similar to the problem in the training process, the inference can also encounter a memory problem. Here we split the input batch into several sub-batches along the image's row dimension (also can be along the column), feed these sub-batches into the decoder one by one, and then assemble them as the harmonization result. Another problem is that in the real world, the resolution of many images is not divisible by the downsampling multiple of the LRIP structure. To deal with that, each block of LRIP takes input vectors of the same batch size (identical to the composite image's size) instead of different ones.

\textbf{Optional 3D LUT prediction.} In the proposed HINet, we can optionally predict the 3D LUT parameters for harmonization. The 3D LUT is a lookup table that maps an RGB value to another value, with which we can harmonize the foreground of the composite image. The motivation of this optional design is for facilitating the manual control of users and enhancing the network's comprehensibility. Since 3D LUT is essentially a global transformation, we predict its parameters by the features for $f_{App}$. Different from the elaborately designed filters in \cite{xue2022dccf, ke2022harmonizer}, we predict the 3D LUT parameters directly and can obtain competitive results. Note again that the 3D LUT is optional and the network performance is not affected by its existence (see \cref{tab:LUT}).

It is also worth noting that the LUT prediction is independent of our INR decoder. That is, users have the flexibility to choose which to use in inference based on their preference for higher harmonization quality (INR decoder) or more control over the result (3D LUT). This is quite different from recent CDTNet \cite{cong2022high} as LUT prediction is an integral part of their final structure.

\section{Experiments}
\label{sec:exp}

\subsection{Experimental Settings}
\label{subsec:exp settings}

\textbf{Datasets.} We follow previous papers to train and evaluate our method on the benchmark dataset iHarmony4 \cite{cong2020dovenet}, which is synthesized using color transformation methods such as \cite{fecker2008histogram} and consists of 4 sub-datasets (HAdobe5k, HCOCO, Hday2night, and HFlickr), with 73146 images in total. For the HR image harmonization, we follow \cite{cong2022high} to evaluate on HAdobe5k sub-dataset that consists of HR images among the four sub-datasets, and also follow \cite{ke2022harmonizer, xue2022dccf} to evaluate on the original resolution iHarmony4 dataset without any downsampling operation. To further illustrate the effectiveness, we follow the existing approaches \cite{hang2022scs, ling2021region, guo2021image, cong2022high} and evaluate our method on 99 LR real composite images released by \cite{tsai2017deep} and 100 HR ones released by \cite{cong2022high}.

\textbf{Evaluation metrics.} We follow the previous methods and evaluate the harmonization performance with Mean Squared Error (MSE), foreground MSE (fMSE, only consider the foreground area), Peak Signal-to-Noise Ratio (PSNR), and Structure Similarity Index Measure (SSIM).

\textbf{Implementation details.}
We adopt the iDIH-HRNet \cite{sofiiuk2021foreground} as the encoder structure. The first three encoder layers are leveraged to predict parameters of $f_{Cont}$ which is a three-block LRIP structure, while the remained layers are utilized to construct a U-Net\cite{ronneberger2015u} like structure and the output is for predicting $f_{App}$ and the optional 3D LUT (See \cref{fig:Pipeline} for more details). If not specified, the number of the three LRIP blocks' hidden layers are 3, 2, and 1 respectively, and we set 2 hidden layers for $f_{App}$. All the hidden layers are of 32 dimensions. We adopt the same positional embedding as \cite{anokhin2021image} and also leverage the Factorized Multiplicative Modulation\cite{skorokhodov2021adversarial} to reduce redundant parameters. The 3D LUT dimension is set to 7.

We only utilize L2 loss to supervise the harmonization results, in addition to an extra regularization ensuring that the 3D LUT values do not overflow. We adopt AdamW\cite{loshchilov2018decoupled} optimizer with an initial learning rate $1e^{-4}$. We train our model for 60 epochs with a batch size of 16, and the learning rate decays in a Cosine Annealing strategy. The model is implemented with Pytorch and we conduct training and evaluation on a single RTX 3090 GPU.

\begin{table*}
\caption{Comparisons on $256\times256$ LR version of the iHarmony4 dataset\cite{cong2020dovenet} with other dense pixel-to-pixel harmonization methods. The best result is shown in bold, while the second best is underlined. The arrow of the metric indicates in which direction the value is better. The results of the state-of-the-art methods are quoted from their source papers, while ``-" indicates that no relevant results are provided.}
    \label{tab:LR SOTA compare}
    \renewcommand{\arraystretch}{1.} 
    \centering
    \resizebox{\textwidth}{!}{
    \begin{tabular}{@{}c|c|ccccccccc|c@{}}
\toprule
Dataset                     & Metric & DIH\cite{tsai2017deep}    & S$^{2}$AM\cite{cun2020improving}   & DoveNet\cite{cong2020dovenet}   & BargainNet\cite{cong2021bargainnet} & RainNet\cite{ling2021region} & IntrinsicIH\cite{guo2021intrinsic} & IHT\cite{guo2021image}   & iDIH-HRNet\cite{sofiiuk2021foreground}      & CDTNet\cite{cong2022high}          & Ours           \\ \midrule
\multirow{3}{*}{HAdobe5K}   & MSE$\downarrow$    & 92.65  & 48.22  & 52.32  & 39.94   & -       & 43.02     & 38.53  & {\ul 21.80}     & \textbf{20.62}  & 23.11          \\
                            & fMSE$\downarrow$   & -      & -      & -      & -       & -       & 284.21    & 265.11 & -              & -               & 170.85         \\
                            & PSNR$\uparrow$   & 32.28  & 35.34  & 34.34  & 35.34   & 36.22   & 35.20      & 36.88  & 37.19          & {\ul 38.24}     & \textbf{38.31} \\ \midrule
\multirow{3}{*}{HCOCO}      & MSE$\downarrow$    & 51.85  & 33.07  & 36.72  & 24.84   & -       & 24.92     & 16.89  & \textbf{13.93} & {\ul 16.25}     & 16.41          \\
                            & fMSE$\downarrow$   & -      & -      & -      & -       & -       & 416.38    & 299.3  & -              & -               & 296.45         \\
                            & PSNR$\uparrow$   & 34.69  & 36.09  & 35.83  & 37.03   & 37.08   & 37.16     & 38.76  & \textbf{39.63} & 39.15           & {\ul 39.16}    \\ \midrule
\multirow{3}{*}{Hday2night} & MSE$\downarrow$    & 82.34  & 48.78  & 54.05  & 50.98   & -       & 55.53     & 53.01  & 60.18          & \textbf{36.72}  & {\ul 51.60}     \\
                            & fMSE$\downarrow$   & -      & -      & -      & -       & -       & 797.04    & 704.42 & -              & -               & 670.32         \\
                            & PSNR$\uparrow$   & 34.62  & 35.60   & 35.18  & 35.67   & 34.83   & 35.96     & 37.10   & 37.71          & \textbf{37.95}  & {\ul 37.81}    \\ \midrule
\multirow{3}{*}{HFlickr}    & MSE$\downarrow$    & 163.38 & 124.53 & 133.14 & 97.32   & -       & 105.13    & 74.51  & \textbf{59.42} & 68.61           & {\ul 68.52}    \\
                            & fMSE$\downarrow$   & -      & -      & -      & -       & -       & 716.60     & 515.45 & -              & -               & 448.77         \\
                            & PSNR$\uparrow$   & 29.55  & 31.00     & 30.21  & 31.34   & 31.64   & 31.34     & 33.13  & \textbf{33.88} & {\ul 33.55}     & 33.53          \\ \midrule
\multirow{3}{*}{All}        & MSE$\downarrow$    & 76.77  & 48.00     & 52.36  & 37.82   & 40.29   & 38.71     & 30.30   & \textbf{22.15} & {\ul 23.75}     & 24.82          \\
                            & fMSE$\downarrow$   & 773.18 & 481.79 & 549.96 & 405.23  & 469.60   & 400.29    & 320.78 & {\ul 256.34}   & \textbf{252.05} & 283.56         \\
                            & PSNR$\uparrow$   & 33.41  & 35.29  & 34.75  & 35.88   & 36.12   & 35.90      & 37.55  & {\ul 38.24}    & 38.23           & \textbf{38.26} \\ \bottomrule
\end{tabular}}
\end{table*}

\begin{figure}[t]
  \centering
   \includegraphics[width=\linewidth]{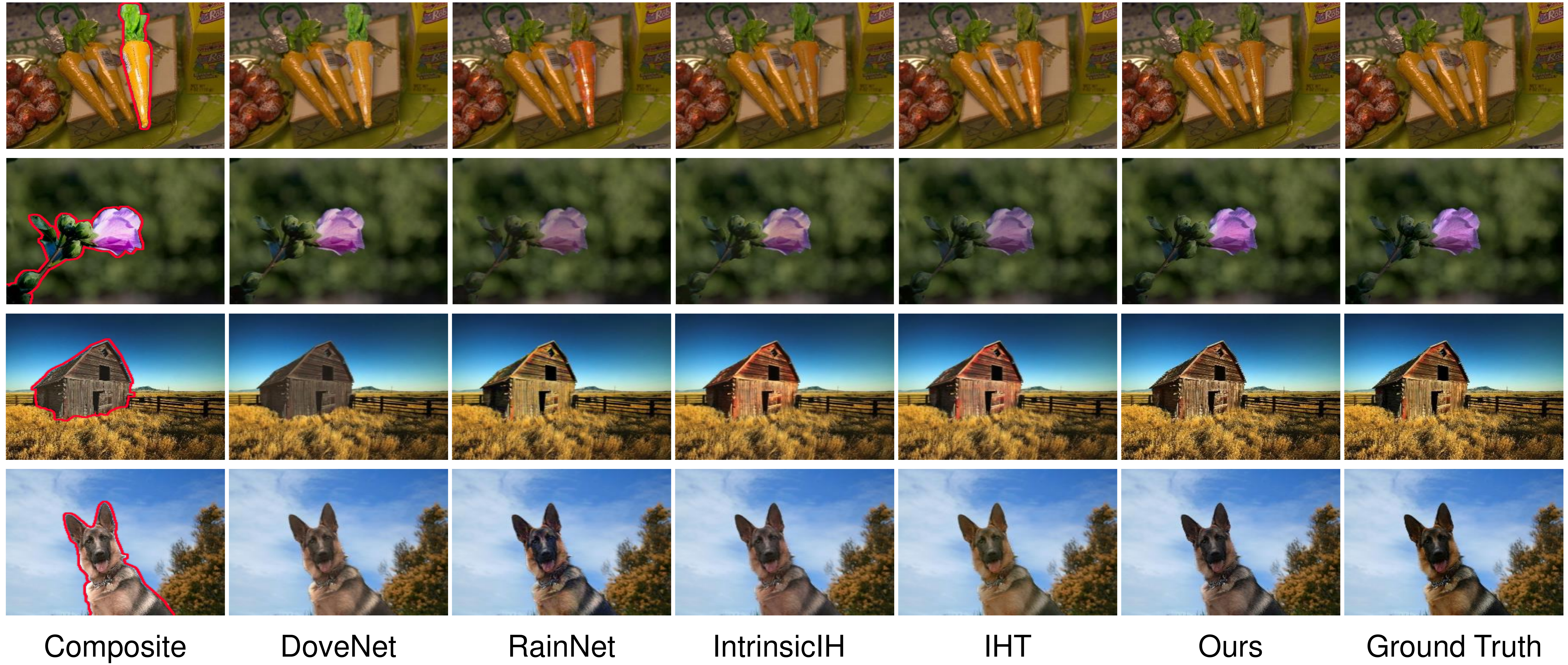}
   \caption{Visual comparisons on $256\times 256$ LR version of iHarmony4 dataset. From left to right, we show the composite images, the results of \cite{cong2020dovenet, ling2021region, guo2021intrinsic, guo2021image} and ours, and the ground truth images. The foreground is stroked by a red line. We have resized the images to their original aspect ratio for a better view.}
   \label{fig:VIS256}
\end{figure}

\begin{figure}[t]
  \centering
   \includegraphics[width=\linewidth]{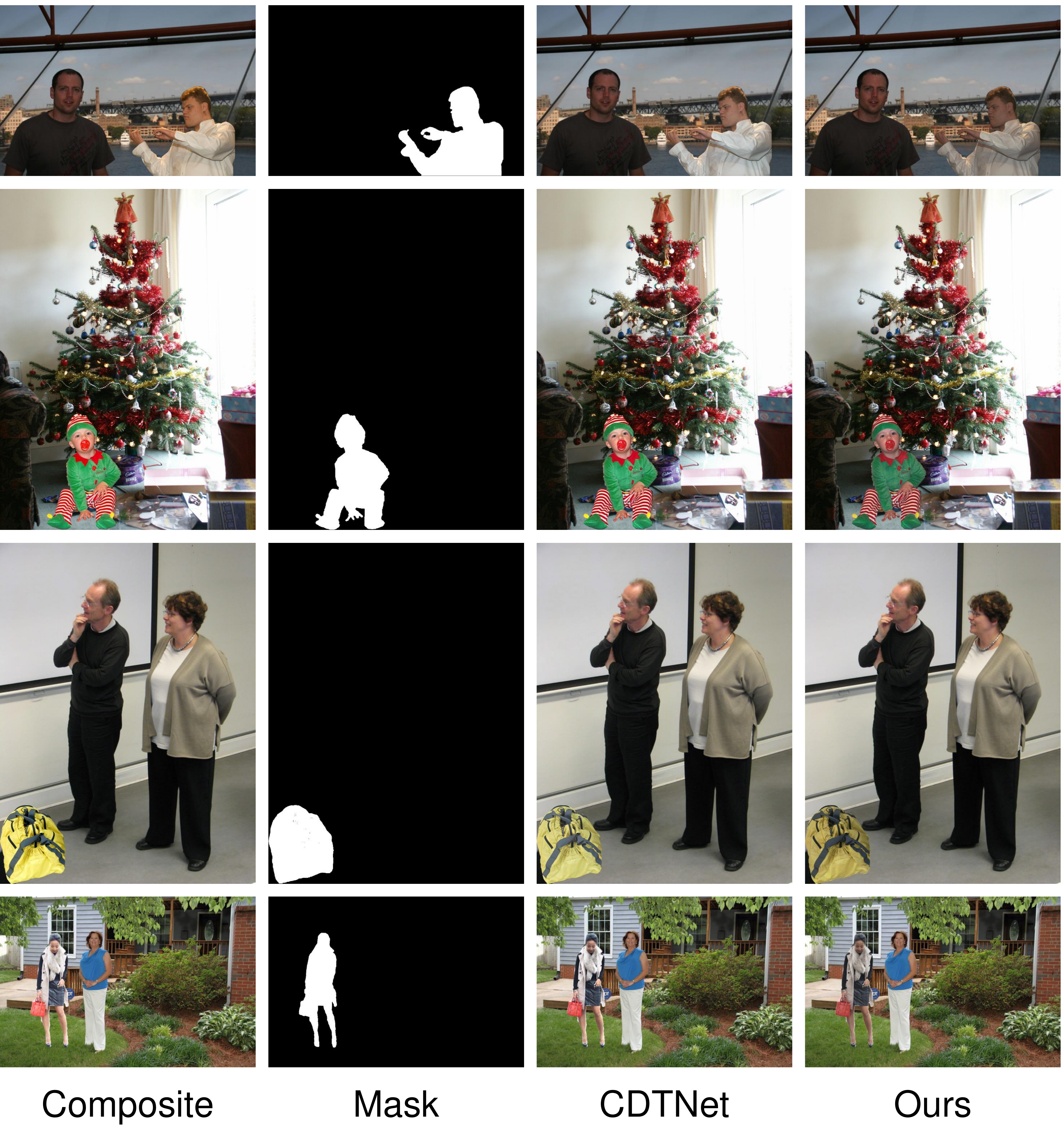}
   \caption{Visual comparisons on 100 HR real composite images ($1024\times1024$) released by \cite{cong2022high}. As there is no ground truth, from left to right, we show the composite images, the masks, and the results of \cite{cong2022high} and ours. We have resized the images to their original aspect ratio for a better view.}
   \label{fig:VISHRReal1024}
\end{figure}

\begin{figure}[t]
  \centering
   \includegraphics[width=\linewidth]{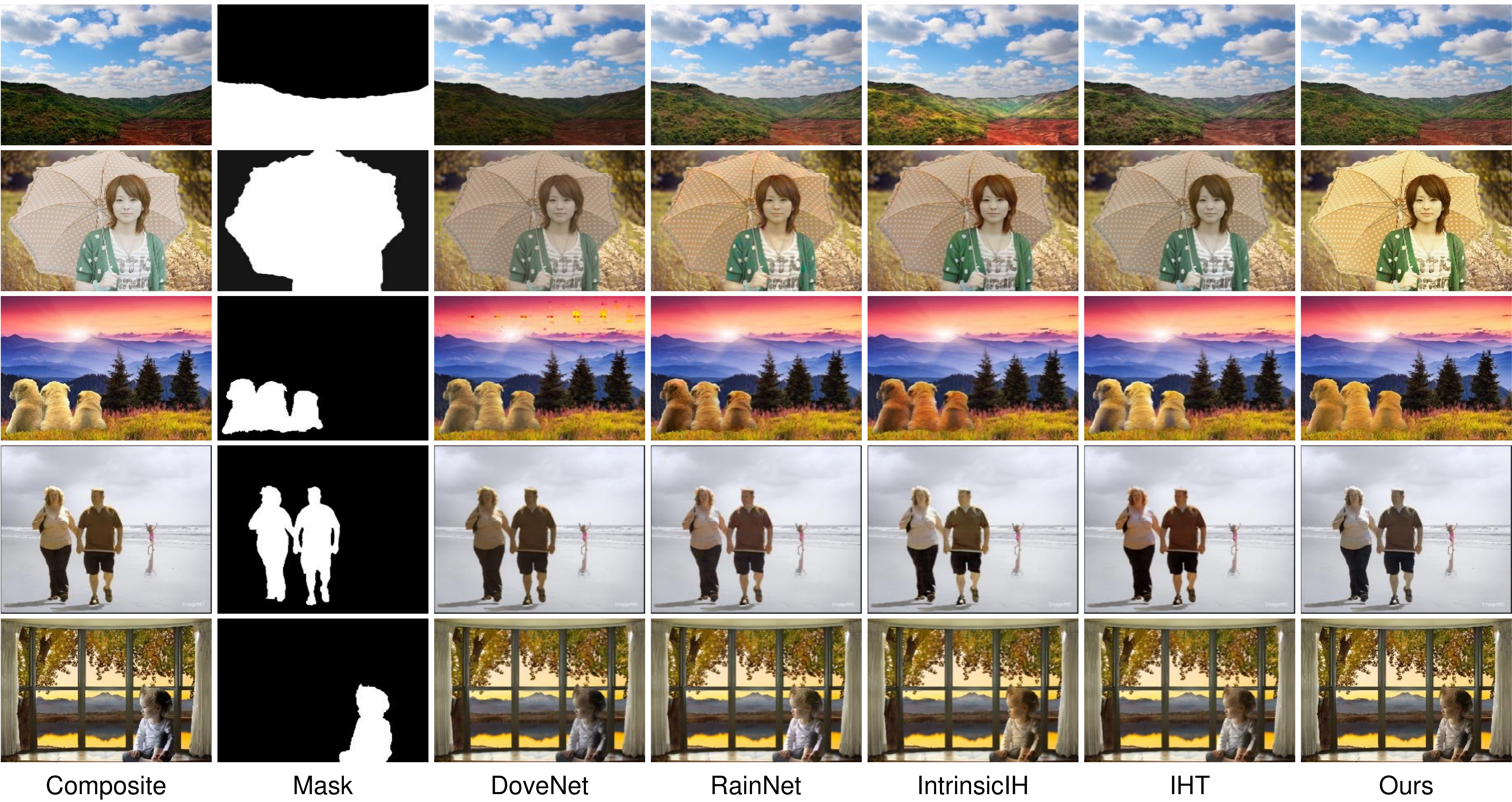}
   \caption{Visual comparisons on 99 LR real composite images ($256\times256$) released by \cite{tsai2017deep}. As there is no ground truth, from left to right, we show the composite images, the masks, and the results of \cite{cong2020dovenet, ling2021region, guo2021intrinsic, guo2021image} and ours. We have resized the images to their original aspect ratio for a better view.}
   \label{fig:VISLRReal256}
\end{figure}

\subsection{Comparison with Existing Methods}
\label{subsec:sota}

\textbf{HR image harmonization.} We here compare our method with the recent HR image harmonization methods \cite{cong2022high, xue2022dccf, ke2022harmonizer, liang2022spatial}, which leverage color-to-color transformations. We conduct experiments on $1024\times1024$ and $2048\times2048$ versions of HAdobe5K sub-dataset, in the same way as in \cite{cong2022high}. Results in \cref{tab:SOTA compare} show that we achieve better performance than \cite{cong2022high} on higher resolution ($2048\times2048$). Although \cite{cong2022high} performs well on $1024\times1024$, with the resolution increasing, its performance drops sharply, while our method maintains stable performance and even achieves better results on the original resolution.

\begin{table*}[!t]
    \caption{User study on 99 LR real composite images released by \cite{tsai2017deep}. For the voting metric, each user can select more than one realistic image. ``Total votes" represents the number of times one method's results are chosen as reality. ``Ratio" denotes the percentage among all votes. For the B-T score, each user must choose the preferred one from a pair of two images. The best value is shown in bold and the second best is underlined.}
    \label{tab:User LR}
    \renewcommand{\arraystretch}{1.2} 
    \centering
\begin{tabular}{c|c|cccccc}
\hline
\multicolumn{2}{c|}{Metric}      & Composite & DoveNet\cite{cong2020dovenet} & RainNet\cite{ling2021region} & IntrinsicIH\cite{guo2021intrinsic}       & IHT\cite{guo2021image}     & Ours          \\ \hline
\multirow{2}{*}{Voting}& Total votes & 161       & 215     & 293     & \textbf{326}     & 299     & {\ul 306}     \\
&Ratio       & 10.06\%   & 13.44\% & 18.31\% & \textbf{20.38\%}  & 18.69\% & {\ul 19.13\%} \\ \hline
\multicolumn{2}{c|}{B-T Score} & 0.0688 & 0.1375 & 0.1848 & 0.1901 & \textbf{0.2285} & {\ul 0.1903}  \\ \hline
\end{tabular}

\end{table*}

\begin{table}[!t]
\caption{User study on 100 HR real composite images released by \cite{cong2022high}. For the voting metric, each user can select more than one realistic image. ``Total votes" represents the number of times one method's results are chosen as reality. ``Ratio" denotes the percentage among all votes. For the B-T score, each user must choose the preferred one from a pair of two images. }
    \label{tab:User HR}
    \renewcommand{\arraystretch}{1.2} 
    \centering
\begin{tabular}{c|c|ccc}
\hline
\multicolumn{2}{c|}{Metric}     & Composite & CDTNet\cite{cong2022high}  & Ours             \\ \hline
\multirow{2}{*}{Voting}& Total votes & 274       & 349     & \textbf{512}     \\
&Ratio       & 24.14\%   & 30.75\% & \textbf{45.11\%} \\ \hline
\multicolumn{2}{c|}{B-T Score} & 0.260 & 0.308 & \textbf{0.432}  \\ \hline
\end{tabular}
    
\end{table}

Following \cite{xue2022dccf, ke2022harmonizer}, we conduct the harmonization experiment on the original resolution of iHarmony4 dataset. It is worth noting that within the four sub-datasets of iHarmony4, HAdobe5K stands out as the sole dataset predominantly comprising images with resolutions ranging from 2K to 6K resolution (1944$\sim$6048), while all other datasets feature images with resolutions below 1024 (HCOCO/120$\sim$640, Hday2night/313$\sim$854, Hflickr/150$\sim$1024). From \cref{tab:SOTA compare}, our method substantially beats the previous DCCF in all metrics (\textit{e.g.}, PSNR 38.67 \textit{vs.} 37.78) on HAdobe5K. This demonstrates our method's effectiveness in handling ultra-HR images and underscores our superiority in real-world harmonization scenarios that frequently involve HR imagery. Besides, we can also observe that the HINet outperforms the two state-of-the-art HR harmonization methods on the entire iHarmony4 dataset.

We also compare with the recent S$^2$CRNet \cite{liang2022spatial}. Since their open-source pre-trained model requires extra semantic labels as input, we do not display the comparison results in \cref{tab:SOTA compare}. Referring to the original paper \cite{liang2022spatial}, both S$^2$CRNet-S and S$^2$CRNet-V get about 36$\sim$37 PSNR on 2048$\times$2048 HAdobe5K, while ours can get 38.35 PSNR (\cref{tab:SOTA compare}). Therefore, our method is still better even without additional label input. We also extend our comparison to more recent works \cite{wang2023semi, guerreiro2023pct}. We cite their results directly from their respective source papers. When evaluated on the 2048$\times$2048 HAdobe5k dataset, our method achieves a PSNR of 38.35, slightly outperforming \cite{wang2023semi} which attains 38.29. While \cite{guerreiro2023pct} achieved notable results with their ViT backbone on the full-resolution iHarmony4 dataset, when evaluated under a similar experimental setup (using a CNN backbone and only L2 loss), our results remained competitive, with our method reaching 38.07 PSNR compared to their 38.05 PSNR.

For visual comparisons, we display the results on $2048\times 2048$ version of HAdobe5K sub-dataset in \cref{fig:VISAdobe2048}, aligned with \cite{cong2022high}, and display the results on the original resolution of iHarmony4 dataset (resolution can reach $6K$) in \cref{fig:VISRAW}, aligned with \cite{xue2022dccf, ke2022harmonizer}.

\textbf{LR image harmonization.} We also evaluate our method on LR image harmonization in \cref{tab:LR SOTA compare}. Since the existing pixel-to-pixel methods \cite{tsai2017deep, cun2020improving, cong2020dovenet, cong2021bargainnet, ling2021region, guo2021intrinsic, guo2021image, sofiiuk2021foreground} cannot be applied to HR images, for fair comparisons, we conduct experiments on $256\times256$ LR iHarmony4 dataset and achieve competitive performance. We also take \cite{cong2022high} into consideration which is a combination of pixel-to-pixel and color-to-color transformations, while we do not consider the recent \cite{hang2022scs} as it introduces extra training data. From the results, we can observe that the HINet can achieve competitive results on LR image harmonization compared with other state-of-the-art methods. Considering the comparison results on HR images displayed in \cref{tab:SOTA compare}, it can be seen that our method achieves more performance gains as the image resolution increases. We visualize the comparisons on $256\times 256$ version of iHarmony4 dataset in \cref{fig:VIS256}.

\textbf{Real composite images.} Since the iHarmony4 is a synthetic dataset \cite{cong2020dovenet}, to better reveal the performance of our method on real images, we follow \cite{cong2022high} to harmonize 100 HR real composite images released by \cite{cong2022high} in \cref{fig:VISHRReal1024} and also follow \cite{guo2021intrinsic, hang2022scs, ling2021region} to visualize the harmonization results on 99 LR real composite images released by \cite{tsai2017deep} in \cref{fig:VISLRReal256}. We conduct user studies for a fair comparison. The result shows the superiority of our method.

Since there are no ground truth images for the real composite images, we cannot leverage the former metrics (PSNR, MSE, and fMSE) to evaluate our performance. Here, we follow \cite{ling2021region} to conduct a user study. Specifically, we invite 8 volunteers to choose the most realistic one/ones from the composite image and the results of the compared methods. Each time, we present an image and its variants to the user. The order is randomly shuffled, thus the users do not know which method the image belongs to. Every volunteer will evaluate the whole 100 HR real composite images and the 99 LR ones. From the results in \cref{tab:User HR} and \cref{tab:User LR}, we can observe that our method can achieve the best performance on HR image harmonization and achieve competitive performance with the state-of-the-art methods on LR image harmonization.

Additionally, we adopted the Bradley-Terry model (B-T model) \cite{bradley1952rank} for ranking, following \cite{cong2020dovenet, ke2022harmonizer}. In this metric, volunteers were presented with pairs of results, randomly sampled from all methods (including composite images). They were required to select the preferred result in each pair. Pairwise comparisons were conducted on the 100 HR real composite images and the 99 LR ones, resulting in 1485 LR pairs and 300 HR pairs. We invited another 5 volunteers to participate in this ranking study, and the corresponding B-T scores are provided in \cref{tab:User HR} and \cref{tab:User LR}. Notably, the conclusions drawn from the aforementioned voting metric remain robust, despite some variations in the rankings of LR images.

\begin{table}[!t]
\caption{Effciency comparison on a single 2048$\times$2048 image.}
\label{tab:Effi}
\renewcommand{\arraystretch}{1.2} 
\centering
\resizebox{1\linewidth}{!}{\begin{tabular}{c|ccc|ccc}
\hline
\multirow{2}{*}{Metrics} & \multicolumn{3}{c|}{Color-to-Color Methods} & \multicolumn{3}{c}{Dense Pixel-to-Pixel Methods} \\ \cline{2-7} 
 & S$^2$CRNet-S & Harmonizer & DCCF & IntrinsicIH & IHT & Ours \\ \hline
Params(M) & 1.15 & 4.73 & 18.09 & 33.80 & 21.80 & 38.21 \\
MACs(G) & 0.605 & 0.036 & 12.677 & OOM & OOM & 36.484 \\
Mem(M) (Train) & 9621 & 4732 & 4159 & OOM & OOM & {\textbf{2835}} \\
Mem(M) (Inference) & 346 & 931 & 1525 & OOM & OOM & 7365/2439/{\textbf{831}} \\
Time(s) & 0.16 & 0.15 & 0.14 & OOM & OOM & {\textbf{0.23}}/0.52/1.12 \\ \hline
\end{tabular}}
\end{table}

\begin{table}[t]
\caption{The efficiency of the design of MLPs decoupling and LRIP structure. The metrics are evaluated on $256\times256$ images with a batch size of 2.}
    \label{tab:Efficiency of LRIP}
    \renewcommand{\arraystretch}{1.3} 
    \centering
    \resizebox{\linewidth}{!}{
\begin{tabular}{c|cc}
\hline
Structure                  & Params (M)     & Mem (MB)         \\ \hline
Pure local MLPs with LRIP & 40.38          & 1233.20          \\
LRIP Blocks with the same size inputs                     & \textbf{38.21} & 1411.16          \\\hline
ours                       & \textbf{38.21} & \textbf{1137.52} \\ \hline
\end{tabular}}
    
\end{table}

\begin{table}[t]
\caption{The efficiency of the design of the RSC training strategy for HR image harmonization. ``Direct finetune" denotes inputting all the vectors for training, not using the RSC strategy.  The metrics are evaluated on images with a batch size of 2.}
    \label{tab:Efficiency of RSC}
    \centering
\begin{tabular}{c|c|c}
\hline
Resolution            & Strategy        & Mem (MB)         \\ \hline
\multirow{2}{*}{$1024\times1024$} & Direct finetune & 4586.35          \\
                      & RSC finetune    & \textbf{4090.86} \\ \hline
\multirow{2}{*}{$2048\times2048$} & Direct finetune & 15641.33         \\
                      & RSC finetune    & \textbf{4090.86} \\ \hline
\end{tabular}
    
\end{table}

\begin{table}[t]
    \caption{Demonstration of the effectiveness of MLPs decoupling. The experiments are conducted on LR iHarmony4 dataset.}
    \label{tab:Decouple}
    \renewcommand{\arraystretch}{1.1} 
    \centering
    \resizebox{0.9\linewidth}{!}{
\begin{tabular}{c|ccc}
\hline
Structure                          & MSE$\downarrow$            & fMSE$\downarrow$            & PSNR$\uparrow$           \\ \hline
Pure global MLPs                   & 25.35          & 290.01          & 38.13          \\
Pure local MLPs                    & 25.56          & 294.97          & 38.00             \\
Decoupled MLPs (ours)              & \textbf{24.82} & \textbf{283.56} & \textbf{38.26} \\ \hline
Params from the last layer             & 25.17               & 288.04                & 38.17      \\
Params from multiple layers (ours) & \textbf{24.82} & \textbf{283.56} & \textbf{38.26} \\ \hline
\end{tabular}}

\end{table}

\begin{table}[t]
    \caption{Demonstration of the effectiveness of the LRIP design. ``No LRIP" denotes only the first block has the input vectors. The experiments are conducted on LR iHarmony4 dataset.}
    \label{tab:LRIP}
    \renewcommand{\arraystretch}{1.1} 
    \centering
    \resizebox{\linewidth}{!}{
\begin{tabular}{c|ccc}
\hline
Structure                      & MSE$\downarrow$            & fMSE$\downarrow$            & PSNR$\uparrow$           \\ \hline
No LRIP                        & 25.65          & 288.41          & 38.10           \\
Blocks with the same batch size inputs & 25.01          & 285.79          & 38.16          \\
LRIP (ours)                    & \textbf{24.82} & \textbf{283.56} & \textbf{38.26} \\ \hline
\end{tabular}}

\end{table}

\begin{table}[t]
\caption{The effectiveness of multiple inputs for HR image harmonization. The experiments are conducted on HAdobe5K sub-dataset. The best result is shown in bold. ``Bilinear resize" denotes interpolating LR harmonization results for HR results, while ``Direct query" denotes querying the decoder that is trained on LR images for HR harmonization results.}
    \label{tab:multiple_input}
    \renewcommand{\arraystretch}{1.2} 
    \centering
    \resizebox{\linewidth}{!}{
\begin{tabular}{c|c|cccc}
\hline
Resolution            & \multicolumn{1}{c|}{Input type \& Strategy} & MSE$\downarrow$            & fMSE$\downarrow$            & PSNR$\uparrow$           & SSIM$\uparrow$            \\ \hline
\multirow{2}{*}{$256\times256$}  & SingleInput                                & 22.79          & 183.85          & 37.92          & 0.9894          \\
                      & MultipleInput                              & \textbf{22.52} & \textbf{182.11} & \textbf{38.00}    & \textbf{0.9897} \\ \hline
\multirow{4}{*}{$1024\times1024$} & SingleInput + Bilinear resize              & 33.84          & 293.11          & 35.69          & 0.9691          \\
                      & MultipleInput + Bilinear resize            & \textbf{32.40}  & \textbf{287.84} & \textbf{35.78} & \textbf{0.9701} \\ \cline{2-6} 
                      & SingleInput + Direct query                 & 69.93          & 631.6           & 32.01          & 0.9346          \\
                      & MultipleInput + Direct query               & \textbf{23.44} & \textbf{194.7}  & \textbf{38.02} & \textbf{0.9883} \\ \hline
\end{tabular}}
    
\end{table}

\begin{table}[!t]
\caption{Demonstration of the effectiveness of RSC strategy on HAdobe5K dataset. ``Direct query" denotes querying the decoder that trained on LR images for HR harmonization results. ``Direct finetune" denotes inputting all the vectors for training, not using the RSC strategy. ``OOM" represents the out-of-memory problem.}
    \label{tab:RSC}
    \renewcommand{\arraystretch}{1.1} 
    \centering
    \resizebox{\linewidth}{!}{
\begin{tabular}{c|c|ccc}
\hline
Resolution                                      & \multicolumn{1}{c|}{Strategy}                       & MSE$\downarrow$            & fMSE$\downarrow$            & PSNR$\uparrow$           \\ \hline
\multirow{4}{*}{$1024\times1024$}                           & Direct query                   & 23.44          & 194.70           & 38.02          \\
                                                & Train from scratch & 36.44          & 263.28          & 36.85          \\
                                                & Direct finetune                & 23.46          & 193.27          & \textbf{38.38} \\
                                                & RSC finetune                   & \textbf{22.68} & \textbf{187.97} & \textbf{38.38} \\ \hline
\multirow{3}{*}{$2048\times2048$}                           & Direct query                   & 26.04          & 212.46          & 37.76          \\
                                                & Direct finetune (OOM)                & \textbf{-}     & \textbf{-}      & -              \\
                                                & RSC finetune                   & \textbf{24.08} & \textbf{192.20}  & \textbf{38.35} \\ \hline
\multirow{3}{*}{\shortstack{Original resolution \\ ($\sim6K$)}} & Direct query                   & 32.63          & 246.33          & 37.07          \\
                                                & Direct finetune (OOM)                & -              & -               & -              \\
                                                & RSC finetune                   & \textbf{21.81} & \textbf{173.72} & \textbf{38.71} \\ \hline
\end{tabular}}
    
\end{table}

\begin{table}[!t]
\caption{Demonstration that the HINet does not rely on 3D LUT prediction. The experiments are conducted on LR iHarmony4.}
    \label{tab:LUT}
    \renewcommand{\arraystretch}{0.9} 
    \centering
    \resizebox{0.9\linewidth}{!}{
\begin{tabular}{c|c|cc}
\hline
\multirow{2}{*}{Metric} & \multirow{2}{*}{Only LUT prediction} & \multicolumn{2}{c}{Extra LUT Prediction}               \\ \cline{3-4} 
                            &                                      & \multicolumn{1}{c|}{w/o}             & w/              \\ \hline
MSE$\downarrow$                         & -                                    & \multicolumn{1}{c|}{\textbf{24.74}}  & 24.82           \\
fMSE$\downarrow$                        & -                                    & \multicolumn{1}{c|}{\textbf{283.05}} & 283.56          \\
PSNR$\uparrow$                        & -                                    & \multicolumn{1}{c|}{\textbf{38.29}}  & 38.26           \\
LUT-MSE$\downarrow$                     & 25.81                                & \multicolumn{1}{c|}{-}               & \textbf{25.56}  \\
LUT-fMSE$\downarrow$                    & 297.72                               & \multicolumn{1}{c|}{-}               & \textbf{293.66} \\
LUT-PSNR$\uparrow$                    & 38.03                                & \multicolumn{1}{c|}{-}               & \textbf{38.08}  \\ \hline
\end{tabular}}

\end{table}

\begin{table}[t]
\caption{Effect of different numbers of channels in MLPs on the final performance. The experiments are conducted on LR iHarmony4 dataset.}
    \label{tab:MLP width}
    \centering
\begin{tabular}{c|ccc}
\hline
MLP width & MSE$\downarrow$                       & fMSE$\downarrow$                       & PSNR$\uparrow$                      \\ \hline
16        & 25.68                     & 293.70                      & 38.04                     \\
32        & \multicolumn{1}{c}{24.82} & \multicolumn{1}{c}{283.56} & \multicolumn{1}{c}{38.26} \\
64        & \textbf{24.74}            & \textbf{283.05}            & \textbf{38.31}            \\ \hline
\end{tabular}
    
\end{table}

\begin{table}[t]
\caption{Effect of different numbers of hidden layers in $f_{Cont}$ on the final performance. The experiments are conducted on LR iHarmony4 dataset.}
    \label{tab:Cont MLP depth}
    \renewcommand{\arraystretch}{1.3} 
    \centering
    \resizebox{0.9\linewidth}{!}{
\begin{tabular}{c|c|ccc}
\hline
$f_{Cont}$ structure                                   & Depth   & MSE$\downarrow$            & fMSE$\downarrow$   & PSNR$\uparrow$           \\ \hline
\multirow{3}{*}{Same depth all blocks}      & 1       & 24.82          & 284.91 & 38.15          \\
                                            & 2       & 25.30          & 288.94 & 38.13          \\
                                            & 3       & 24.89          & 282.58 & 38.24          \\ \hline
\multirow{2}{*}{Different depth each block} & 1, 2, 3 & \textbf{24.26} & \textbf{282.00} & 38.24          \\
                                            & 3, 2, 1 & 24.82          & 283.56 & \textbf{38.26} \\ \hline
\end{tabular}}
    
\end{table}

\begin{table}[!ht]
\caption{Effect of different numbers of hidden layers in $f_{App}$ on the final performance. The experiments are conducted on LR iHarmony4 dataset.}
    \label{tab:App MLP depth}
    \centering
\begin{tabular}{c|ccc}
\hline
$f_{App}$ depth & \multicolumn{1}{c}{MSE$\downarrow$} & \multicolumn{1}{c}{fMSE$\downarrow$} & \multicolumn{1}{c}{PSNR$\uparrow$} \\ \hline
1     & 24.91                   & 286.64                   & 38.23                    \\
2     & 24.82                   & \textbf{283.56}          & \textbf{38.26}           \\
4     & \textbf{24.62}          & 286.73                   & 38.17                    \\ \hline
\end{tabular}
    
\end{table}

\begin{table}[!ht]
\caption{Performance of different position embeddings. The experiments are conducted on LR iHarmony4 dataset.}
    \label{tab:Positonal embeddings}
    \centering
\begin{tabular}{c|ccc}
\hline
Positional embeddings & MSE$\downarrow$            & fMSE$\downarrow$            & PSNR$\uparrow$           \\ \hline
Nerf\cite{mildenhall2021nerf}                  & 33.64          & 321.59          & 37.95          \\
RFF\cite{tancik2020fourier}               & 28.15          & 326.94          & 37.62          \\
INR-GAN\cite{skorokhodov2021adversarial}                & 25.48          & 285.64          & \textbf{38.28} \\
CIPS\cite{anokhin2021image}                  & \textbf{24.82} & \textbf{283.56} & 38.26          \\ \hline
\end{tabular}
    
\end{table}

\begin{table}[!ht]
\caption{Performance of different 3D LUT dimensions. The experiments are conducted on LR iHarmony4 dataset.}
    \label{tab:3D LUT dimensions}
    \centering
    \resizebox{\linewidth}{!}{
\begin{tabular}{c|ccc}
\hline
3D LUT dimensions & LUT-MSE$\downarrow$            & LUT-fMSE$\downarrow$            & LUT-PSNR$\uparrow$           \\ \hline
5                 &\textbf{25.55} &299.47 &37.95          \\
7                 & 25.81 & \textbf{297.72} & \textbf{38.03} \\
13                & 25.59          & 298.48          & 37.93          \\
17                & 25.90          & 305.43          & 37.88          \\ \hline
\end{tabular}}
    
\end{table}

\begin{figure}[t]
  \centering
   \includegraphics[width=\linewidth]{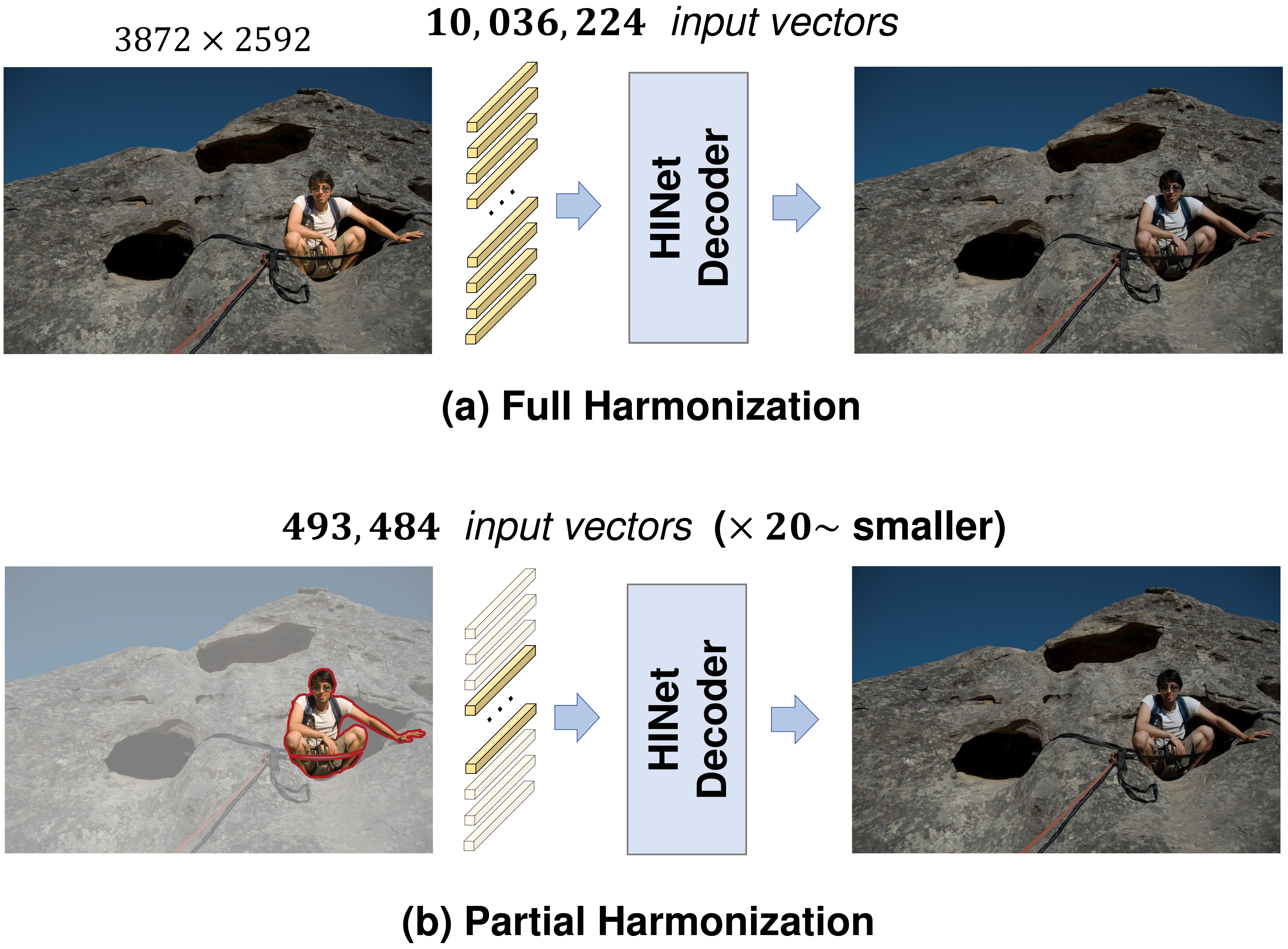}
   \caption{Region-based harmonization of composite images. (a) displays the normal harmonization process that feeds all the vectors into the decoder, while (b) displays the region-based harmonization process.}
   \label{fig:Partial_Harmonization_Image}
\end{figure}

\subsection{Efficiency Analyses}
\label{subsec:efficiency}

In \cref{tab:Effi}, we take a single 2048$\times$2048 image as an example and compare the efficiency with existing methods from the perspective of model parameters (Params), calculation amount (MACs), memory overhead (Mem) during training and inference, and inference runtime (Time). For the memory cost and runtime of inference, we show results when the input is split into 1/4/16 parts (see \cref{subsec:specific}). From the results, thanks to the RSC strategy designed in \cref{subsec:specific}, we have the lowest training memory cost (less than 3GB) among all methods. By varying the number of input splits, we can achieve competitive performance with color-to-color methods either on inference memory cost or runtime. Regarding model parameters, our approach offers flexibility during the inference phase. We can prune it to exclusively employ our INR decoder (34.71M parameters) or solely utilize the predicted 3D LUT (23.98M parameters), adapting to different scenarios. Consequently, we can match the parameter count with other dense pixel-to-pixel methods. Most importantly, we hope to draw the attention that our method is the first dense pixel-to-pixel method that can handle HR images ($\sim$6K, others encounter out-of-memory (OOM)). Considering pixel-to-pixel methods can model more complex scenarios than color-to-color ones (see \cref{sec:relate}), our method, with SOTA harmonization performance and competitive efficiency, is of great significance and beneficial to this field's development.

Moreover. to validate the efficiency of our design, we here adopt Model Parameters (Params) and GPU Memory Cost (Mem) metrics. From \cref{tab:Efficiency of LRIP}, we can see that splitting MLPs into $f_{Cont}$ and $f_{App}$ can both reduce parameters and save memory cost compared with the structure of pure local MLPs, while the design of LRIP that leverages different batch sizes of input vectors can save much memory. From \cref{tab:Efficiency of RSC}, we can observe that, with image resolution increasing, the memory cost of the direct finetuning grows sharply, while there is no impact on the memory cost of our RSC strategy, nor is the performance (Please see \cref{tab:LUT}).

\subsection{Ablation Studies}
\label{subsec:ablation}

\textbf{Effectiveness of MLPs decoupling.} To verify the effectiveness of the decoupled design of $f_{Cont}$ and $f_{App}$, we conduct experiments in \cref{tab:Decouple} where we modify the HINet with pure global MLPs or local MLPs. We also compare with a modified version that the parameters of MLPs are all from the last layer of the encoder. The results have validated the superiority of our design.

\textbf{Effectiveness of LRIP structure.} We compare with the HINet without LRIP structure in \cref{tab:LRIP}, from which we can observe that the network with the LRIP structure achieves higher accuracy. To go a step further, we also compare with the structure with each block fed inputs of the same batch size (LRIP is fed inputs of gradually increasing batch size), and the LRIP again achieves better results.

\textbf{Evaluation of specific designs.} As mentioned in \cref{subsec:specific}, we leverage multiple inputs to help harmonize HR images. To compare with only using the grid coordinate as input, experiments are conducted as shown in \cref{tab:multiple_input}. We can observe that although different types of inputs have competitive results on LR image harmonization, when applied to HR images, only the coordinate input cannot achieve satisfying accuracy, even worse than using bilinear interpolation (only resize the foreground, while the other parts keep the same as the composite image).

To evaluate the effectiveness of the designed RSC HR training strategy, we conduct experiments in \cref{tab:RSC} on the HAdobe5K sub-dataset, as it has much higher resolution among the four datasets in the iHarmony4 dataset\cite{cong2022high}. We compare with results by directly querying the decoder trained on LR images for HR harmonization, and results by directly finetuning/training from scratch the network with HR images (here we only consider $1024\times1024$, images with higher resolution will encounter an out-of-memory problem). From the results, we can see that our proposed RSC training strategy greatly improves the performance of HR image harmonization.

In \cref{tab:LUT}, we show that using an additional 3D LUT prediction head has trivial impacts on the final harmonization results (even has a 0.03 PSNR drop). Thus we do not rely on it like \cite{cong2022high, xue2022dccf, ke2022harmonizer}. Moreover, we can observe that with our method, the predicted 3D LUT can achieve better results.

\textbf{Settings of MLPs width.} We here test the effect of different numbers of channels in MLPs on the harmonization performance. From \cref{tab:MLP width}, we can observe that with the MLPs' width increasing, the results get better. Considering the practical memory limitation, we choose 32 channels as the final design.

\textbf{Settings of MLPs depth.} In this part, we test different settings of the MLP hidden layers' number in $f_{Cont}$ and $f_{App}$ respectively. From the results in \cref{tab:Cont MLP depth}, the effectiveness of adopting decreasing number of hidden layers for blocks in the LRIP structure is verified. Not only does it maintain harmonization performance, but such a structure also saves a lot of memory cost (the batch size of the input vectors increases from the first block to the last block). From the results in \cref{tab:App MLP depth}, we set two hidden layers for $f_{App}$.

\textbf{Choices of positional embedding.} As mentioned in \cite{mildenhall2021nerf, tancik2020fourier}, a proper positional embedding is important for INR performance. Here we try several different positional embeddings\cite{mildenhall2021nerf, tancik2020fourier, anokhin2021image, skorokhodov2021adversarial}. From the results in \cref{tab:Positonal embeddings}, we finally adopt the positional embedding in \cite{anokhin2021image}.

\textbf{3D LUT dimensions.} To evaluate the effect of different 3D LUT dimensions, we conduct experiments in \cref{tab:3D LUT dimensions}. We can observe that there is a peak in the middle. Dimensions that are too large or too small can both lead to performance degradation. We consider that small dimensions may be not sufficient enough to cover the variation range, while large dimensions may cause redundant parameters' predictions, leading to much burden on the encoder.

\begin{figure*}[t]
  \centering
   \includegraphics[width=\textwidth]{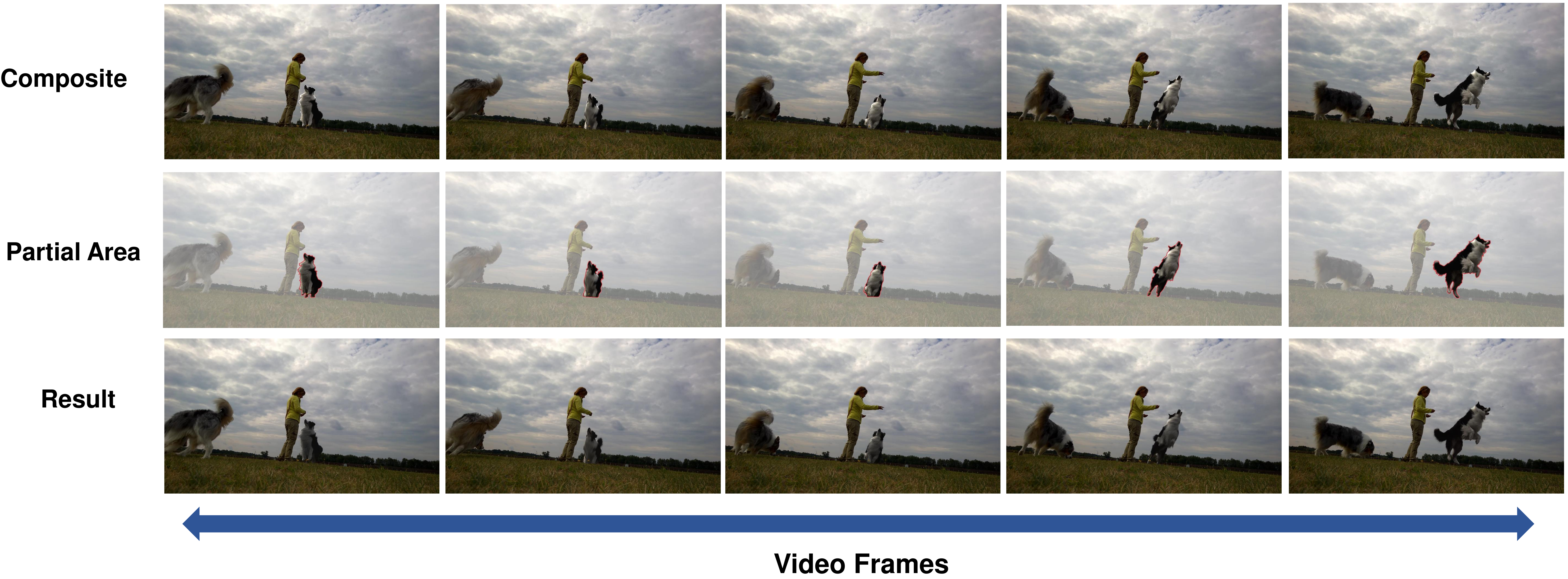}
   \caption{Region-based harmonization of composite video clips. The first row denotes the composite frames. In the second row, we stroke the partial area whose vectors are fed into the decoder with a red line, while the remained untouched region is made transparent. We display our harmonization results in the third row. Please zoom in for a better review.}
   \label{fig:Partial_Harmonization_Video}
\end{figure*}

\section{Applications}
\label{sec:usage}

\subsection{Region-based image harmonization}
As mentioned in \cref{subsec:specific}, one property of the INR decoder is that the input is no longer a feature map but a batch of vectors. By leveraging this feature, we can achieve region-based harmonization by only feeding the vectors inside the foreground area into the decoder, while leaving the remained area untouched. In this way, the proposed HINet can harmonize a partial area of the composite image, thus saving much memory cost and achieving speedup. This is a feature that existing methods\cite{cong2022high, sofiiuk2021foreground, hang2022scs} do not have, since their CNN decoder needs to receive the output features of the entire image from the encoder as input, while we utilize the INR paradigm whose input is a batch of vectors $V$. \cref{fig:Partial_Harmonization_Image} displays the region-based harmonization process. Given a $3K$ composite image, if we feed the vectors of the entire image into the decoder (somewhat similar to the CNN process), there will be about 10 million input vectors, which will cause much memory consumption and slow processing speed. While, if we apply region-based harmonization and only harmonize the vectors in the foreground region, the number of input vectors will be reduced to 0.5 million, just equivalent to processing half a $1K$ image!

Take advantage of this potential, we can even apply the HINet to video harmonization, while preventing much memory cost. We here just take the video as a stack of image frames, and the process is similar to \cref{fig:Partial_Harmonization_Image}. We display the video harmonization process in \cref{fig:Partial_Harmonization_Video}.

\begin{figure}[t]
  \centering
   \includegraphics[width=\linewidth]{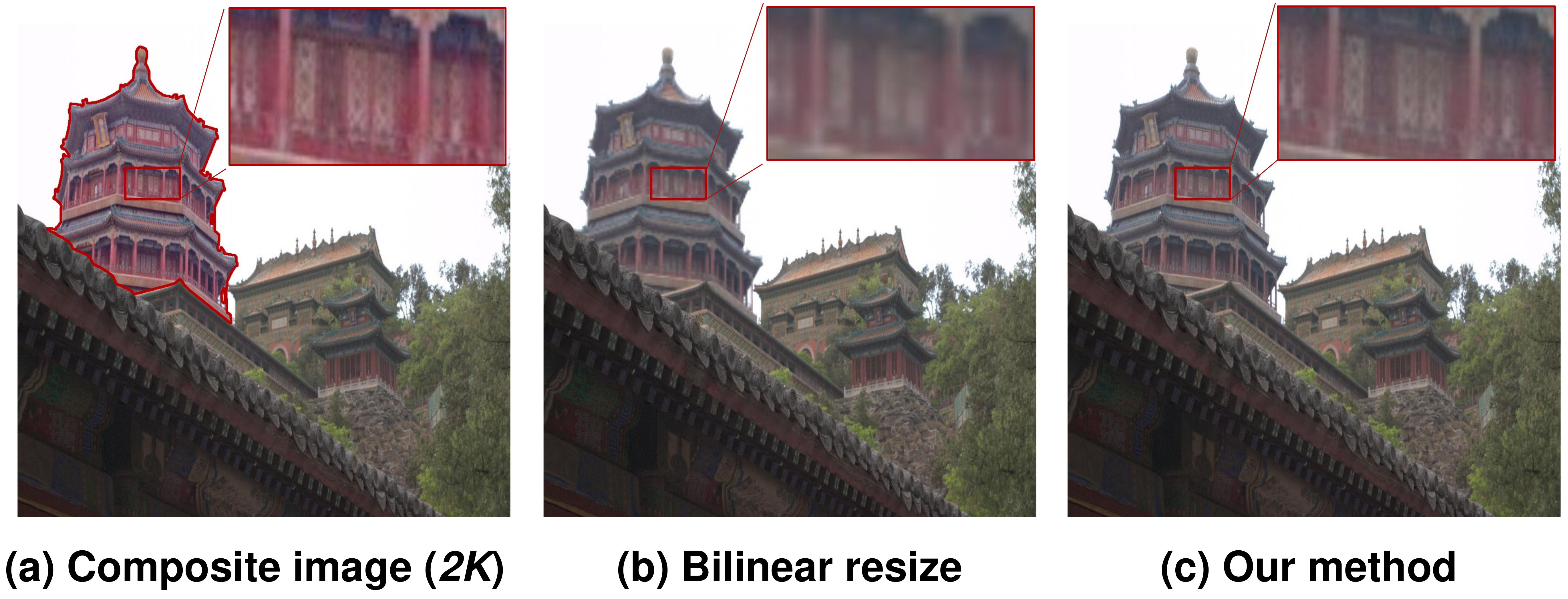}
   \caption{Arbitrary resolution harmonization potential of our method. Take a $2K$ image (a) as an example, the harmonization result by bilinear interpolation (b) losses much information, while ours (c) keeps high fidelity even though we only train the network on LR images (Please zoom in for a better view).}
   \label{fig:Arbitrary_resolution}
\end{figure}

\begin{figure}[t]
  \centering
   \includegraphics[width=\linewidth]{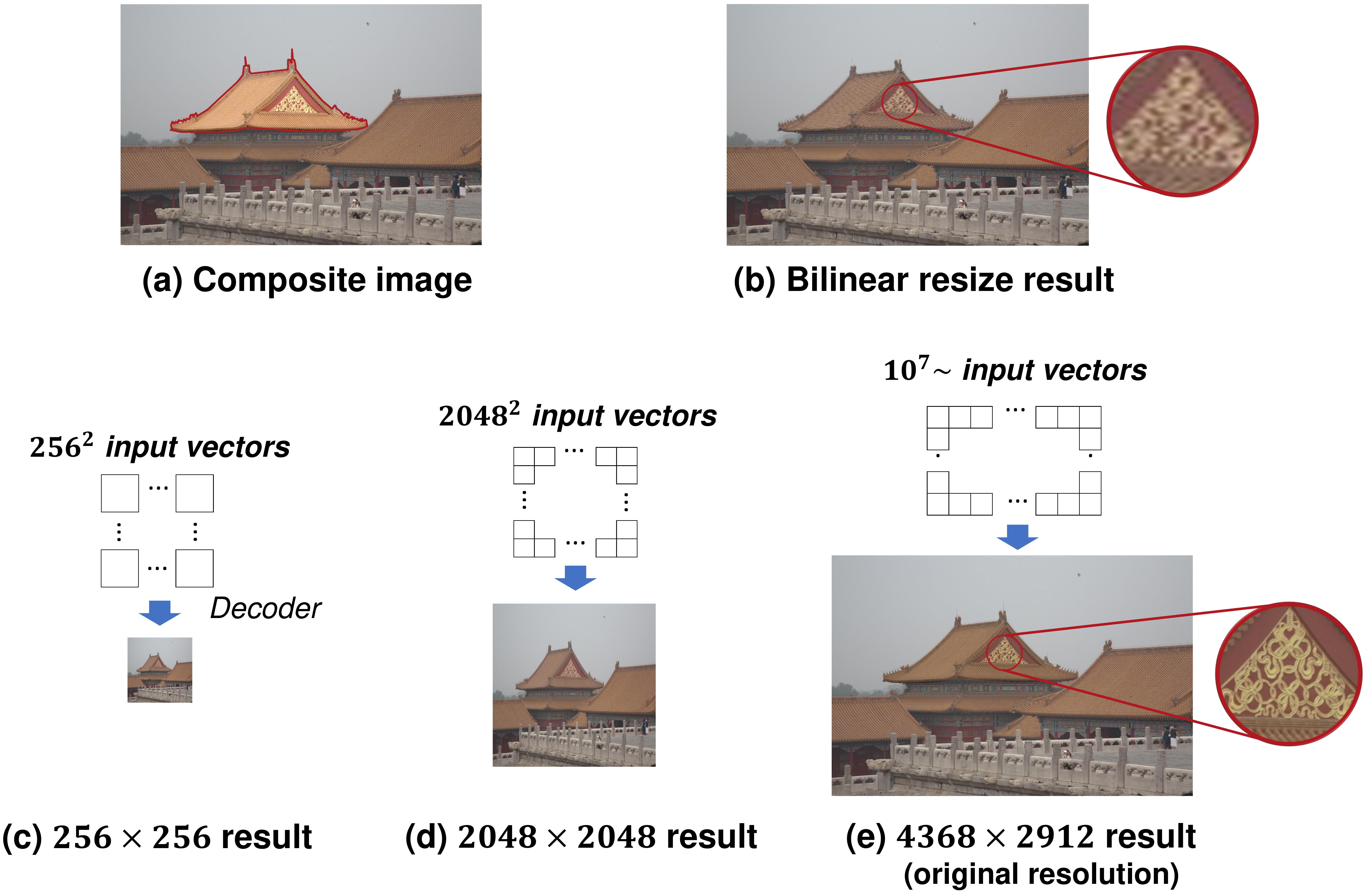}
   \caption{Illustration of arbitrary resolution harmonization. (a) is the composite image with the foreground stroked by a red line. (c), (d), and (e) are harmonization results with different resolutions. (b) is the result of bilinearly resizing (c) (only resize the foreground, other region keeps the same as (a)). Please zoom in for a better view.}
   \label{fig:Details_Arbitrary_resolution}
\end{figure}

\begin{figure}[t]
  \centering
   \includegraphics[width=\linewidth]{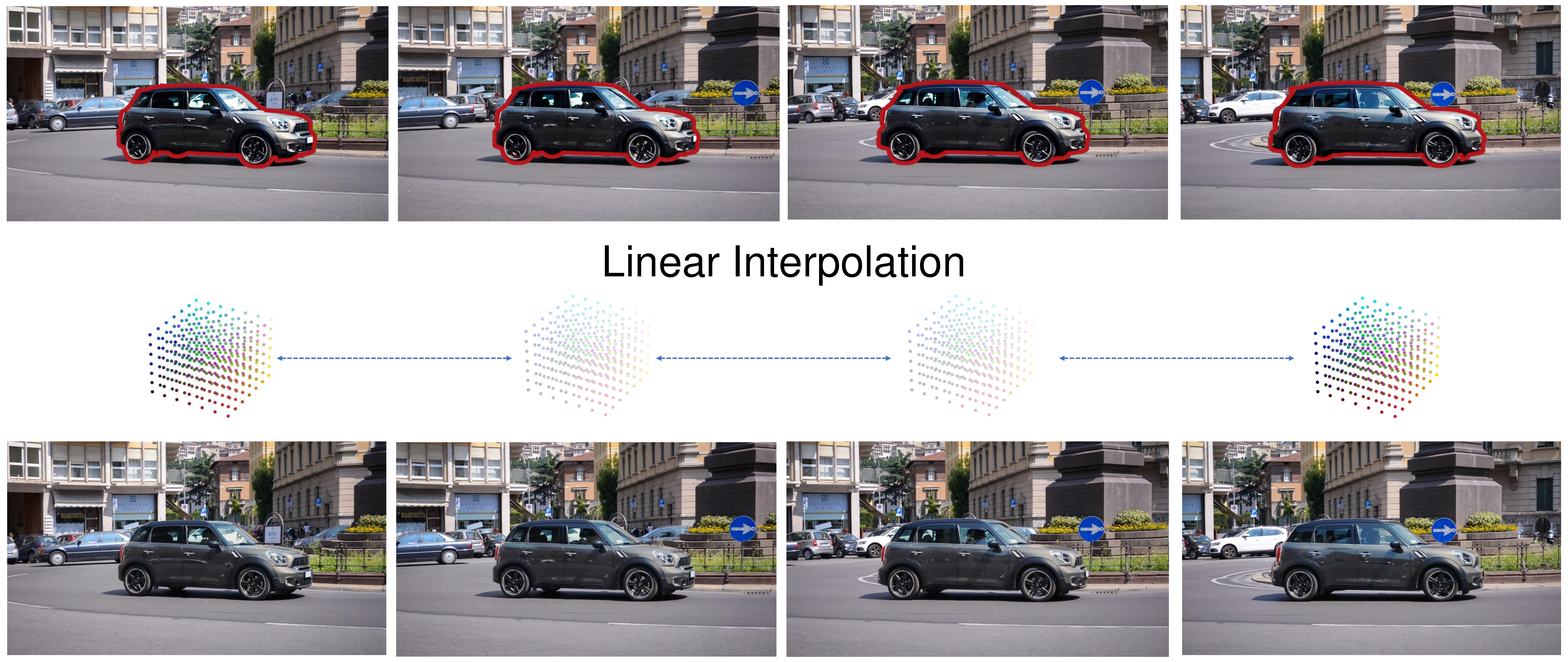}
   \caption{Illustration of smooth 3D LUT interpolation on video harmonization. The first row is frames of the composite video with the foreground stroked by a red line. The second row is the predicted 3D LUT. The third row is the harmonization result with the interpolated 3D LUT. Please zoom in for a better view.}
   \label{fig:3DLUT_linear}
\end{figure}

\subsection{Arbitrary resolution image harmonization}
One defect of the previous methods \cite{guo2021intrinsic, hang2022scs} is that once the network structures are configured, these methods can only harmonize the images with a fixed resolution. If we want to harmonize images with other resolutions, the network structure must be re-configured and retrained. To demonstrate it more clearly, suppose that we have a pure convolution structure designed for $256\times 256$ images, then when the input is $256\times 256$ or a multiple of that, the network can handle it. However, if the input is \eg $256\times 257$ (only one value increased), then the network may fail to process it, as the resolution $257$ is not divisible by the downsampling multiple of the network's encoder. It is feasible to resize the image to the nearest multiple of $256\times 256$, but since the downsampling multiple of the encoder is usually large\cite{sofiiuk2021foreground}, much information may be lost if the image is directly resized to a lower resolution (See \cref{fig:Arbitrary_resolution} as an example), while if the image is resized to a higher resolution, then there will be much redundant computational overhead. As a comparison, the proposed HINet can achieve arbitrary resolution image harmonization. Since our method is built based on the INR paradigm, we consider the task as harmonizing continuous images rather than discrete image arrays. Therefore, in our method, we only need to sample intermediate coordinates to construct an input batch with the size identical to the image, and then can produce a higher fidelity harmonization result than direct using interpolation, even if the target resolution has never been seen by the model (See \cref{fig:Arbitrary_resolution} and the experiments in \cref{tab:multiple_input}). We illustrate this feature in \cref{fig:Details_Arbitrary_resolution} for better comprehensibility. Given vectors of different batch sizes, we can get harmonization results at different resolutions.

\subsection{Optional usage of 3D LUT}
\label{subsec: application 3d}
In \cref{subsec:ablation}, we mention that the HINet can optionally predict a 3D LUT for controllable harmonization, and the existence of this optional part will not affect the final performance of the network. The motivation follows \cite{xue2022dccf, ke2022harmonizer} to make more space for manual control of the harmonization result, since the HINet is essentially a black box with little control by the user. Different from the complex design of the hand-crafted image filters in \cite{xue2022dccf, ke2022harmonizer}, we adopt 3D LUT, which is a global RGB-to-RGB mapping. Although \cite{ke2022harmonizer} claimed that directly predicting the filters' parameters cannot achieve good results, for which they utilized a hierarchical structure, in the HINet, we have no special design for 3D LUT prediction but directly predict it by the encoder's features, and the performance still looks good (Please refer to \cref{tab:LUT}).

The potential usages of 3D LUT are two-fold. On the one hand, the users can better understand how the network harmonizes composite images. Along with that, they can manually change the harmonization result by modifying the parameters of the 3D LUT, which is easy with the help of PhotoShop and 3D LUT Creator tools. On the other hand, if high harmonization quality is not pursued, we can apply 3D LUTs to video harmonization for fast and continuous results. As mentioned in \cite{ke2022harmonizer}, harmonizing each video frame independently can lead to flickering phenomena, so they leverage the exponential moving average on the hand-crafted image filters' arguments for smoothness. Inspired by them, we here linearly interpolate the parameters of 3D LUT for continuous results. Suppose the 3D LUT parameters at $M_{th}$ frame is $lut_M$, that at $N_{th}$ frame is $lut_N$, then the parameters at intermediate frames $(M:N)$ can be formulated as:
\begin{equation}
  lut_{K} = lut_M + (lut_N - lut_M) \times \frac{K - M}{N - M}, K \in (M: N)
  \label{eq:lut_linear}
\end{equation}
Then, we can extract several frames at intervals from the video, predict their 3D LUT, and interpolate the other frames'. It allows for video harmonization to be fairly quick and results to be continuous. We display the process in \cref{fig:3DLUT_linear}.

\begin{table}[!ht]
\caption{Performance on video harmonization. The experiments are conducted on the HYouTube dataset. ``TL" denotes temporal loss. The best result is shown in bold, while the second best is underlined.}
    \label{tab:video harmonization}
    \centering
\begin{tabular}{@{}cccccc@{}}
\toprule
Method & fMSE & MSE & PSNR & TL & Time(s) \\ \midrule
\multicolumn{1}{c|}{Lu \etal \cite{hyoutube2021}} & 186.72 & 26.50 & 37.61 & \textbf{5.11} & 2.37 \\
\multicolumn{1}{c|}{Harmonizer} & 211.49 & 32.28 & 36.48 & 23.72 & 1.49 \\
\multicolumn{1}{c|}{Harmonizer (EMA)} & 197.36 & 30.17 & 36.84 & 17.11 & 1.49 \\ \midrule
\multicolumn{1}{c|}{Ours(LUT)} & 176.99 & 24.45 & 38.56 & 7.09 & {\ul 1.23} \\
\multicolumn{1}{c|}{Ours(LUT-Interpolation)} & {\ul 171.80} & {\ul 23.73} & {\ul 38.71} & 6.73 & \textbf{0.86} \\
\multicolumn{1}{c|}{Ours(Decoder)} & \textbf{159.42} & \textbf{22.38} & \textbf{39.12} & {\ul 6.06} & 1.53 \\ \bottomrule
\end{tabular}
\end{table}

\subsection{Video Harmonization}
For a comprehensive evaluation, we extended our method to the video harmonization task and conducted experiments on the publicly available HYouTube dataset \cite{hyoutube2021}, consisting of 3194 video samples, each consisting of 20 frames. We resized the frames to 256$\times$256, in alignment with \cite{hyoutube2021}. We compared our approach with Harmonizer \cite{ke2022harmonizer} and the video harmonization framework presented in \cite{hyoutube2021}. Both Harmonizer and our method were trained from scratch on the training set of HYouTube. Regarding \cite{hyoutube2021}, we directly reference the results reported in their source paper. As Harmonizer is originally designed for image harmonization, we also compare its exponential-moving-average (EMA) variant tailored for video harmonization. We set its EMA coefficient to 1/6, corresponding to a frame interval of 167ms in HYouTube, which is also in line with Harmonizer's source code. For our method, we assess both the performance of the INR decoder and the 3D LUT. We also evaluated the 3D LUT interpolation strategy mentioned in \cref{subsec: application 3d}, where we sample five key frames from the 20-frame video clip. We measured performance using metrics such as fMSE, MSE, and PSNR. Additionally, we employed the Temporal Loss (TL) in \cite{hyoutube2021} to assess temporal consistency and measured the inference time for processing a 20-frame video clip on a single RTX 3090 with the batch size set to 1.

The results, presented in \cref{tab:video harmonization}, indicate that our method outperforms the other methods in almost every metric for video harmonization, except for temporal consistency. This demonstrates the superior generalization of our approach. Regarding temporal consistency, our 3D LUT interpolation strategy not only enhances performance but also accelerates the harmonization process. In contrast, Harmonizer's EMA strategy, while reducing the TL metric value, does not improve processing speed since it still calculates every frame's result. Both Harmonizer and our method fall short when compared to \cite{hyoutube2021}. We attribute this to the fact that \cite{hyoutube2021} incorporates temporal information, considering several previous and future frames in their network's input, thereby having access to more temporal data. In contrast, Harmonizer and our approach rely on simpler averaging and interpolation strategies.

\section{Limitations}
\label{sec:Limitations}

Although we have carefully designed the structure, the limitation of the INR still remains, especially on the speed performance when being applied to ultra-HR image harmonization. Since we need to split the input of the decoder into different parts to avoid being out of memory, the memory performance improves along with a certain speed sacrifice. Moreover, compared with the existing methods that leverage a U-Net like structure, there is still space to better fuse the shallow features and deep features in the HINet, which can provide richer features for the MLPs' predictions. We leave these limitations to our future work.

\section{Conclusion}
\label{sec:conclusion}

In this paper, we explore a novel method for HR image harmonization with dense pixel-to-pixel transformations. We leverage the implicit neural representation and carefully design the decoder's structure to ensure visual harmony and reasonable memory cost. To our best knowledge, the proposed HINet is the first dense pixel-to-pixel harmonization method that can be applied to images $\sim6K$ without any hand-crafted image filter and is also the first approach that leverages INR for the harmonization task. Experiments conducted on iHarmony4 dataset have demonstrated the effectiveness of our method for HR image harmonization. Some application potentials in practical usage are explored. We expect that our work can pave way for more research on deep learning-based HR image harmonization.


\normalem
\bibliographystyle{IEEEtran}  
\bibliography{IEEEabrv,Harmonization}

\begin{thebibliography}{10}
\providecommand{\url}[1]{#1}
\csname url@samestyle\endcsname
\providecommand{\newblock}{\relax}
\providecommand{\bibinfo}[2]{#2}
\providecommand{\BIBentrySTDinterwordspacing}{\spaceskip=0pt\relax}
\providecommand{\BIBentryALTinterwordstretchfactor}{4}
\providecommand{\BIBentryALTinterwordspacing}{\spaceskip=\fontdimen2\font plus
\BIBentryALTinterwordstretchfactor\fontdimen3\font minus \fontdimen4\font\relax}
\providecommand{\BIBforeignlanguage}[2]{{%
\expandafter\ifx\csname l@#1\endcsname\relax
\typeout{** WARNING: IEEEtran.bst: No hyphenation pattern has been}%
\typeout{** loaded for the language `#1'. Using the pattern for}%
\typeout{** the default language instead.}%
\else
\language=\csname l@#1\endcsname
\fi
#2}}
\providecommand{\BIBdecl}{\relax}
\BIBdecl

\bibitem{10.1145/1201775.882269}
\BIBentryALTinterwordspacing
P.~P\'{e}rez, M.~Gangnet, and A.~Blake, ``Poisson image editing,'' in \emph{ACM SIGGRAPH 2003 Papers}, ser. SIGGRAPH '03.\hskip 1em plus 0.5em minus 0.4em\relax New York, NY, USA: Association for Computing Machinery, 2003, p. 313–318. [Online]. Available: \url{https://doi.org/10.1145/1201775.882269}
\BIBentrySTDinterwordspacing

\bibitem{barnes2009patchmatch}
C.~Barnes, E.~Shechtman, A.~Finkelstein, and D.~B. Goldman, ``Patchmatch: A randomized correspondence algorithm for structural image editing,'' \emph{ACM Trans. Graph.}, vol.~28, no.~3, p.~24, 2009.

\bibitem{kwatra2003graphcut}
V.~Kwatra, A.~Sch{\"o}dl, I.~Essa, G.~Turk, and A.~Bobick, ``Graphcut textures: Image and video synthesis using graph cuts,'' \emph{ACM Trans. Graph.}, vol.~22, no.~3, pp. 277--286, 2003.

\bibitem{yun2019cutmix}
S.~Yun, D.~Han, S.~J. Oh, S.~Chun, J.~Choe, and Y.~Yoo, ``Cutmix: Regularization strategy to train strong classifiers with localizable features,'' in \emph{Proceedings of the IEEE/CVF international conference on computer vision}, 2019, pp. 6023--6032.

\bibitem{zhang2020learning}
L.~Zhang, T.~Wen, J.~Min, J.~Wang, D.~Han, and J.~Shi, ``Learning object placement by inpainting for compositional data augmentation,'' in \emph{European Conference on Computer Vision}.\hskip 1em plus 0.5em minus 0.4em\relax Springer, 2020, pp. 566--581.

\bibitem{wang2020constrained}
H.~Wang, Q.~Wang, H.~Zhang, J.~Yang, and W.~Zuo, ``Constrained online cut-paste for object detection,'' \emph{IEEE Transactions on Circuits and Systems for Video Technology}, vol.~31, no.~10, pp. 4071--4083, 2020.

\bibitem{wang2023composited}
Y.~Wang, L.~Tang, Y.~Zhong, and B.~Li, ``From composited to real-world: Transformer-based natural image matting,'' \emph{IEEE Transactions on Circuits and Systems for Video Technology}, 2023.

\bibitem{inoue2020learning}
N.~Inoue and T.~Yamasaki, ``Learning from synthetic shadows for shadow detection and removal,'' \emph{IEEE Transactions on Circuits and Systems for Video Technology}, vol.~31, no.~11, pp. 4187--4197, 2020.

\bibitem{ren2018lecarm}
Y.~Ren, Z.~Ying, T.~H. Li, and G.~Li, ``Lecarm: Low-light image enhancement using the camera response model,'' \emph{IEEE Transactions on Circuits and Systems for Video Technology}, vol.~29, no.~4, pp. 968--981, 2018.

\bibitem{zhao2021retinexdip}
Z.~Zhao, B.~Xiong, L.~Wang, Q.~Ou, L.~Yu, and F.~Kuang, ``Retinexdip: A unified deep framework for low-light image enhancement,'' \emph{IEEE Transactions on Circuits and Systems for Video Technology}, vol.~32, no.~3, pp. 1076--1088, 2021.

\bibitem{tsai2017deep}
Y.-H. Tsai, X.~Shen, Z.~Lin, K.~Sunkavalli, X.~Lu, and M.-H. Yang, ``Deep image harmonization,'' in \emph{Proceedings of the IEEE Conference on Computer Vision and Pattern Recognition}, 2017, pp. 3789--3797.

\bibitem{cong2020dovenet}
W.~Cong, J.~Zhang, L.~Niu, L.~Liu, Z.~Ling, W.~Li, and L.~Zhang, ``Dovenet: Deep image harmonization via domain verification,'' in \emph{Proceedings of the IEEE/CVF Conference on Computer Vision and Pattern Recognition}, 2020, pp. 8394--8403.

\bibitem{guo2021image}
Z.~Guo, D.~Guo, H.~Zheng, Z.~Gu, B.~Zheng, and J.~Dong, ``Image harmonization with transformer,'' in \emph{Proceedings of the IEEE/CVF International Conference on Computer Vision}, 2021, pp. 14\,870--14\,879.

\bibitem{ling2021region}
J.~Ling, H.~Xue, L.~Song, R.~Xie, and X.~Gu, ``Region-aware adaptive instance normalization for image harmonization,'' in \emph{Proceedings of the IEEE/CVF Conference on Computer Vision and Pattern Recognition}, 2021, pp. 9361--9370.

\bibitem{guo2021intrinsic}
Z.~Guo, H.~Zheng, Y.~Jiang, Z.~Gu, and B.~Zheng, ``Intrinsic image harmonization,'' in \emph{Proceedings of the IEEE/CVF Conference on Computer Vision and Pattern Recognition}, 2021, pp. 16\,367--16\,376.

\bibitem{sofiiuk2021foreground}
K.~Sofiiuk, P.~Popenova, and A.~Konushin, ``Foreground-aware semantic representations for image harmonization,'' in \emph{Proceedings of the IEEE/CVF Winter Conference on Applications of Computer Vision}, 2021, pp. 1620--1629.

\bibitem{hang2022scs}
Y.~Hang, B.~Xia, W.~Yang, and Q.~Liao, ``Scs-co: Self-consistent style contrastive learning for image harmonization,'' in \emph{Proceedings of the IEEE/CVF Conference on Computer Vision and Pattern Recognition}, 2022, pp. 19\,710--19\,719.

\bibitem{lalonde2007using}
J.-F. Lalonde and A.~A. Efros, ``Using color compatibility for assessing image realism,'' in \emph{2007 IEEE 11th International Conference on Computer Vision}.\hskip 1em plus 0.5em minus 0.4em\relax IEEE, 2007, pp. 1--8.

\bibitem{xue2012understanding}
S.~Xue, A.~Agarwala, J.~Dorsey, and H.~Rushmeier, ``Understanding and improving the realism of image composites,'' \emph{ACM Transactions on graphics (TOG)}, vol.~31, no.~4, pp. 1--10, 2012.

\bibitem{reinhard2001color}
E.~Reinhard, M.~Adhikhmin, B.~Gooch, and P.~Shirley, ``Color transfer between images,'' \emph{IEEE Computer graphics and applications}, vol.~21, no.~5, pp. 34--41, 2001.

\bibitem{pitie2005n}
F.~Pitie, A.~C. Kokaram, and R.~Dahyot, ``N-dimensional probability density function transfer and its application to color transfer,'' in \emph{Tenth IEEE International Conference on Computer Vision (ICCV'05) Volume 1}, vol.~2.\hskip 1em plus 0.5em minus 0.4em\relax IEEE, 2005, pp. 1434--1439.

\bibitem{ronneberger2015u}
O.~Ronneberger, P.~Fischer, and T.~Brox, ``U-net: Convolutional networks for biomedical image segmentation,'' in \emph{International Conference on Medical image computing and computer-assisted intervention}.\hskip 1em plus 0.5em minus 0.4em\relax Springer, 2015, pp. 234--241.

\bibitem{cong2022high}
W.~Cong, X.~Tao, L.~Niu, J.~Liang, X.~Gao, Q.~Sun, and L.~Zhang, ``High-resolution image harmonization via collaborative dual transformations,'' in \emph{Proceedings of the IEEE/CVF Conference on Computer Vision and Pattern Recognition}, 2022, pp. 18\,470--18\,479.

\bibitem{ke2022harmonizer}
Z.~Ke, C.~Sun, L.~Zhu, K.~Xu, and R.~W. Lau, ``Harmonizer: Learning to perform white-box image and video harmonization,'' in \emph{European Conference on Computer Vision}.\hskip 1em plus 0.5em minus 0.4em\relax Springer, 2022, pp. 690--706.

\bibitem{xue2022dccf}
B.~Xue, S.~Ran, Q.~Chen, R.~Jia, B.~Zhao, and B.~Zhao, ``Dccf: Deep comprehensible color filter learning framework for high-resolution image harmonization,'' in \emph{European Conference on Computer Vision}, 2022.

\bibitem{sitzmann2020implicit}
V.~Sitzmann, J.~Martel, A.~Bergman, D.~Lindell, and G.~Wetzstein, ``Implicit neural representations with periodic activation functions,'' \emph{Advances in Neural Information Processing Systems}, vol.~33, pp. 7462--7473, 2020.

\bibitem{tancik2020fourier}
M.~Tancik, P.~Srinivasan, B.~Mildenhall, S.~Fridovich-Keil, N.~Raghavan, U.~Singhal, R.~Ramamoorthi, J.~Barron, and R.~Ng, ``Fourier features let networks learn high frequency functions in low dimensional domains,'' \emph{Advances in Neural Information Processing Systems}, vol.~33, pp. 7537--7547, 2020.

\bibitem{mildenhall2021nerf}
B.~Mildenhall, P.~P. Srinivasan, M.~Tancik, J.~T. Barron, R.~Ramamoorthi, and R.~Ng, ``Nerf: Representing scenes as neural radiance fields for view synthesis,'' \emph{Communications of the ACM}, vol.~65, no.~1, pp. 99--106, 2021.

\bibitem{land1971lightness}
E.~H. Land and J.~J. McCann, ``Lightness and retinex theory,'' \emph{Josa}, vol.~61, no.~1, pp. 1--11, 1971.

\bibitem{land1977retinex}
E.~H. Land, ``The retinex theory of color vision,'' \emph{Scientific american}, vol. 237, no.~6, pp. 108--129, 1977.

\bibitem{karaimer2016software}
H.~C. Karaimer and M.~S. Brown, ``A software platform for manipulating the camera imaging pipeline,'' in \emph{European Conference on Computer Vision}.\hskip 1em plus 0.5em minus 0.4em\relax Springer, 2016, pp. 429--444.

\bibitem{zeng2020learning}
H.~Zeng, J.~Cai, L.~Li, Z.~Cao, and L.~Zhang, ``Learning image-adaptive 3d lookup tables for high performance photo enhancement in real-time,'' \emph{IEEE Transactions on Pattern Analysis and Machine Intelligence}, 2020.

\bibitem{wang2020deep}
J.~Wang, K.~Sun, T.~Cheng, B.~Jiang, C.~Deng, Y.~Zhao, D.~Liu, Y.~Mu, M.~Tan, X.~Wang \emph{et~al.}, ``Deep high-resolution representation learning for visual recognition,'' \emph{IEEE transactions on pattern analysis and machine intelligence}, vol.~43, no.~10, pp. 3349--3364, 2020.

\bibitem{sunkavalli2010multi}
K.~Sunkavalli, M.~K. Johnson, W.~Matusik, and H.~Pfister, ``Multi-scale image harmonization,'' \emph{ACM Transactions on Graphics (TOG)}, vol.~29, no.~4, pp. 1--10, 2010.

\bibitem{cun2020improving}
X.~Cun and C.-M. Pun, ``Improving the harmony of the composite image by spatial-separated attention module,'' \emph{IEEE Transactions on Image Processing}, vol.~29, pp. 4759--4771, 2020.

\bibitem{huang2017arbitrary}
X.~Huang and S.~Belongie, ``Arbitrary style transfer in real-time with adaptive instance normalization,'' in \emph{Proceedings of the IEEE international conference on computer vision}, 2017, pp. 1501--1510.

\bibitem{liang2022spatial}
J.~Liang, X.~Cun, C.-M. Pun, and J.~Wang, ``Spatial-separated curve rendering network for efficient and high-resolution image harmonization,'' in \emph{European Conference on Computer Vision}.\hskip 1em plus 0.5em minus 0.4em\relax Springer, 2022, pp. 334--349.

\bibitem{wang2023semi}
K.~Wang, M.~Gharbi, H.~Zhang, Z.~Xia, and E.~Shechtman, ``Semi-supervised parametric real-world image harmonization,'' in \emph{Proceedings of the IEEE/CVF Conference on Computer Vision and Pattern Recognition}, 2023, pp. 5927--5936.

\bibitem{guerreiro2023pct}
J.~J.~A. Guerreiro, M.~Nakazawa, and B.~Stenger, ``Pct-net: Full resolution image harmonization using pixel-wise color transformations,'' in \emph{Proceedings of the IEEE/CVF Conference on Computer Vision and Pattern Recognition}, 2023, pp. 5917--5926.

\bibitem{stanley2007compositional}
K.~O. Stanley, ``Compositional pattern producing networks: A novel abstraction of development,'' \emph{Genetic programming and evolvable machines}, vol.~8, no.~2, pp. 131--162, 2007.

\bibitem{park2019deepsdf}
J.~J. Park, P.~Florence, J.~Straub, R.~Newcombe, and S.~Lovegrove, ``Deepsdf: Learning continuous signed distance functions for shape representation,'' in \emph{Proceedings of the IEEE/CVF conference on computer vision and pattern recognition}, 2019, pp. 165--174.

\bibitem{mescheder2019occupancy}
L.~Mescheder, M.~Oechsle, M.~Niemeyer, S.~Nowozin, and A.~Geiger, ``Occupancy networks: Learning 3d reconstruction in function space,'' in \emph{Proceedings of the IEEE/CVF conference on computer vision and pattern recognition}, 2019, pp. 4460--4470.

\bibitem{choy20163d}
C.~B. Choy, D.~Xu, J.~Gwak, K.~Chen, and S.~Savarese, ``3d-r2n2: A unified approach for single and multi-view 3d object reconstruction,'' in \emph{European conference on computer vision}.\hskip 1em plus 0.5em minus 0.4em\relax Springer, 2016, pp. 628--644.

\bibitem{fan2017point}
H.~Fan, H.~Su, and L.~J. Guibas, ``A point set generation network for 3d object reconstruction from a single image,'' in \emph{Proceedings of the IEEE conference on computer vision and pattern recognition}, 2017, pp. 605--613.

\bibitem{wang2018pixel2mesh}
N.~Wang, Y.~Zhang, Z.~Li, Y.~Fu, W.~Liu, and Y.-G. Jiang, ``Pixel2mesh: Generating 3d mesh models from single rgb images,'' in \emph{Proceedings of the European conference on computer vision (ECCV)}, 2018, pp. 52--67.

\bibitem{wang2021lossy}
J.~Wang, H.~Zhu, H.~Liu, and Z.~Ma, ``Lossy point cloud geometry compression via end-to-end learning,'' \emph{IEEE Transactions on Circuits and Systems for Video Technology}, vol.~31, no.~12, pp. 4909--4923, 2021.

\bibitem{benedek2016lidar}
C.~Benedek, B.~G{\'a}lai, B.~Nagy, and Z.~Jank{\'o}, ``Lidar-based gait analysis and activity recognition in a 4d surveillance system,'' \emph{IEEE Transactions on Circuits and Systems for Video Technology}, vol.~28, no.~1, pp. 101--113, 2016.

\bibitem{shaham2021spatially}
T.~R. Shaham, M.~Gharbi, R.~Zhang, E.~Shechtman, and T.~Michaeli, ``Spatially-adaptive pixelwise networks for fast image translation,'' in \emph{Proceedings of the IEEE/CVF Conference on Computer Vision and Pattern Recognition}, 2021, pp. 14\,882--14\,891.

\bibitem{chen2021learning}
Y.~Chen, S.~Liu, and X.~Wang, ``Learning continuous image representation with local implicit image function,'' in \emph{Proceedings of the IEEE/CVF conference on computer vision and pattern recognition}, 2021, pp. 8628--8638.

\bibitem{anokhin2021image}
I.~Anokhin, K.~Demochkin, T.~Khakhulin, G.~Sterkin, V.~Lempitsky, and D.~Korzhenkov, ``Image generators with conditionally-independent pixel synthesis,'' in \emph{Proceedings of the IEEE/CVF Conference on Computer Vision and Pattern Recognition}, 2021, pp. 14\,278--14\,287.

\bibitem{skorokhodov2021adversarial}
I.~Skorokhodov, S.~Ignatyev, and M.~Elhoseiny, ``Adversarial generation of continuous images,'' in \emph{Proceedings of the IEEE/CVF Conference on Computer Vision and Pattern Recognition}, 2021, pp. 10\,753--10\,764.

\bibitem{karras2021alias}
T.~Karras, M.~Aittala, S.~Laine, E.~H{\"a}rk{\"o}nen, J.~Hellsten, J.~Lehtinen, and T.~Aila, ``Alias-free generative adversarial networks,'' \emph{Advances in Neural Information Processing Systems}, vol.~34, pp. 852--863, 2021.

\bibitem{fecker2008histogram}
U.~Fecker, M.~Barkowsky, and A.~Kaup, ``Histogram-based prefiltering for luminance and chrominance compensation of multiview video,'' \emph{IEEE Transactions on Circuits and Systems for Video Technology}, vol.~18, no.~9, pp. 1258--1267, 2008.

\bibitem{loshchilov2018decoupled}
I.~Loshchilov and F.~Hutter, ``Decoupled weight decay regularization,'' in \emph{International Conference on Learning Representations}, 2018.

\bibitem{cong2021bargainnet}
W.~Cong, L.~Niu, J.~Zhang, J.~Liang, and L.~Zhang, ``Bargainnet: Background-guided domain translation for image harmonization,'' in \emph{2021 IEEE International Conference on Multimedia and Expo (ICME)}.\hskip 1em plus 0.5em minus 0.4em\relax IEEE, 2021, pp. 1--6.

\bibitem{bradley1952rank}
R.~A. Bradley and M.~E. Terry, ``Rank analysis of incomplete block designs: I. the method of paired comparisons,'' \emph{Biometrika}, vol.~39, no. 3/4, pp. 324--345, 1952.

\bibitem{hyoutube2021}
X.~Lu, S.~Huang, L.~Niu, W.~Cong, and L.~Zhang, ``Deep video harmonization with color mapping consistency,'' \emph{IJCAI}, 2022.

\end{thebibliography}

\end{document}